\pdfoutput=1

\documentclass[11pt]{article}

\usepackage{EMNLP2023}

\usepackage{times}
\usepackage{latexsym}

\usepackage[T1]{fontenc}

\usepackage[utf8]{inputenc}

\usepackage{microtype}

\usepackage{inconsolata}

\usepackage{tabularx}
\usepackage{soul}
\usepackage{graphicx}
\usepackage{enumitem}
\usepackage{colortbl}
\usepackage{tabu}
\usepackage{booktabs}
\usepackage{tabularx}
\usepackage{ragged2e}
\usepackage{fancyhdr}
\fancypagestyle{firstpage}
{
    \fancyhead[L]{}    
    \fancyhead[R]{This paper is accepted for publication at EMNLP 2023. The official version of record is in the ACL Anthology.}
    \fancyfoot{}
}

\definecolor{singlechoice}{RGB}{214,39,40} 
\definecolor{multichoice}{RGB}{44,160,44} 
\definecolor{freetext}{RGB}{31,119,180} 

\usepackage{cleveref}
\crefname{section}{\S}{\S}
\crefname{Section}{\S}{\S}
\crefformat{section}{§#2#1#3}
\crefname{table}{Tab.}{Tab.}
\crefname{appendix_table}{Tab.}{Tab.}
\crefname{Table}{Tab.}{Tab.}
\crefname{Figure}{Fig.}{Fig.}
\crefname{figure}{Fig.}{Fig.}
\crefname{appendix}{Appendix}{Appendix}
\crefname{chapter}{Chapter}{Chapter}

\title{The Past, Present and Better Future of Feedback Learning in Large Language Models for Subjective Human Preferences and Values}

\author{Hannah Rose Kirk$^{1\ddagger}$, Andrew M. Bean$^{1}$, Bertie Vidgen$^{1}$, Paul Röttger$^{2}$, Scott A. Hale$^{1, 3}$\\
  { $^1$University of Oxford, $^2$Bocconi University, $^3$Meedan} \\
  { $^\ddagger$hannah.kirk@oii.ox.ac.uk}\\
}

\begin{document}
\maketitle
\thispagestyle{firstpage}
\begin{abstract}
Human feedback is increasingly used to steer the behaviours of Large Language Models (LLMs). However, it is unclear how to collect and incorporate feedback in a way that is efficient, effective and unbiased, especially for highly subjective human preferences and values.
In this paper, we survey existing approaches for learning from human feedback, drawing on 95 papers primarily from the ACL and arXiv repositories.
First, we summarise \textbf{the past}, pre-LLM trends for integrating human feedback into language models.
Second, we give an overview of \textbf{present} techniques and practices, as well as the motivations for using feedback; conceptual frameworks for defining values and preferences; and how feedback is collected and from whom. Finally, we encourage a \textbf{better future} of feedback learning in LLMs by raising five unresolved conceptual and practical challenges.

\end{abstract}

\section{Introduction}
Incorporating human feedback into Large Language Models (LLMs) is a welcome development to create models that are better aligned with human preferences or values, and exhibit traits such as helpfulness, honesty and harmlessness \cite{askellGeneral2021, baiTraining2022} or safety, quality and groundedness \cite{thoppilanLaMDA2022}. However, learning from human feedback introduces new biases and challenges, and there are many unresolved questions in this fast-moving field of research.  
It is important to take stock of current practices, possible blindspots, and new frontiers of research, so that tangible progress can continue to be made. In this paper, we adopt the dual aim to both survey existing literature on human feedback learning, then draw on the regularities, commonalities and critiques of this survey to also provide recommendations for future work.
We review 95 articles that use human feedback to steer, guide or tailor the behaviours of language models.
This includes making models' responses more coherent and engaging \cite{luBoosting2022}; assisting models to infer user intent \cite{ouyangTraining2022}; rejecting and rebutting unsafe requests \cite{ganguliRed2022, baiTraining2022}; or minimising the risk of hallucination \cite{nakanoWebGPT2021, glaeseImproving2022}. We source articles primarily from the ACL and arXiv repositories, coding each according to a detailed conceptual and methodological schema. 
Our review makes three contributions:
\begin{itemize}[nolistsep]
\item \textbf{The Past}: We include articles released both before and after the advent of LLMs, which avoids recency bias and allows us to track developments through time.
\item \textbf{The Present}: We summarise current practices for incorporating human feedback learning into LLMs, such as reinforcement learning fine-tuning, supervised fine-tuning, and pre-training. We also document how feedback is collected and from whom.
\item \textbf{The Future}: We draw on the findings of our review to highlight five unresolved challenges in the field; two challenges are conceptual and three are practical. 
Conceptual challenges revolve around the fundamental difficulty of specifying a clear shared set of preferences and values. And, even if the conceptual challenges can be resolved, practical challenges remain for converting abstract concepts into reliable signals to guide model behaviours.
\end{itemize}

We find that current processes of incorporating human feedback into LLMs often rely on unsatisfactory simplifying assumptions about the stability, universality and objectivity of human preferences and values. 
What counts as a ``good'', ``high-quality'', ``preferred'' or ``value-aligned'' output is only objective in the abstract \cite{kirkEmpty2023b}; so, we explicitly centre our review on subjective human preferences and values because we believe most text attributes retain some degree of contextual subjectivity. With this in mind, we call for more open, democratically-grounded and interdisciplinary collaboration, supported by robust processes of external scrutiny, to decide how different voices shape current and future LLMs.

\begin{table*}[h]
  \centering
  \small
  \renewcommand{\arraystretch}{1.75}
  \begin{tabularx}{\textwidth}{|>{\hsize=0.5\hsize}X|>{\hsize=0.5\hsize}X|}
    \hline
    \textbf{Keywords} & \textbf{Stemmed Keywords} \\
    \hline
    alignment, human, \hl{value}, moral, ethic, feedback, \newline reinforcement, instruction, red teaming, red-teaming, \newline preferences, harm, honest, \hl{helpful}, \hl{personalis}, \hl{personaliz} &
    align, human, \hl{value}, moral, ethic, feedback, reinforc, instruct, red team, red-team, prefer, harm, honest, \hl{helpful}, \hl{personalis}, \hl{personaliz} \\
    \hline
  \end{tabularx}
  \caption{Keywords for retrieving articles from ACL and arXiv repositories. Highlighted keywords were not stemmed due to irrelevant matches e.g., ``value'' as ``valu'' returns many false positives including the title word ``evaluation''.}
  \label{tab:keywords_tab}
\end{table*}
\section{Methods}
\subsection{Selecting Articles}
\label{sec:selecting}
We use a semi-automated method, casting a wide net of keywords to retrieve articles, then manually assessing their relevance for our review (see \cref{tab:keywords_tab} for keywords and \cref{sec:flowchart} for a schema).

\paragraph{Initial set ($S_0$)} We retrieve articles from two corpora. First, we download the ACL anthology as a \texttt{.bib} file. Second, we use the arXiv API with the computation and language subclass (cs.CL) to find new or industry-led preprints that are not peer-reviewed but have early impact on the field. We match titles with $\ge2$ keywords ($n=187$), and deduplicate dual-posted articles ($n=175$).\footnote{We match on title to increase relevancy and because some ACL articles lack abstract metadata. In a sample of 100 retrieved articles, we determined that $\ge2$ keywords best balanced relevancy with the size of the retrieved set. The cut-off for our automated collection is 17/02/2023.}

\paragraph{Inclusion criteria} Two authors read the abstract and introduction of $S_0$ articles, and included them if all the following questions were answered `yes':
\begin{enumerate}[nolistsep]
    \item \textbf{Topic}: \textit{Does the article seek alignment or adaptation of AI systems to human preferences and values?} This criterion excludes articles that functionally align aspects of language models e.g., word embedding alignment or sequence alignment.
    \item \textbf{Modality}: \textit{Does the article discuss language agents or language as its primary modality?} This criterion excludes any multimodal models, delegate agents or games-based RL.
    \item \textbf{Empirical}: \textit{Does the article contain empirical analysis or artefacts (such as experiments, datasets or models)?} This criterion excludes opinion papers, reviews or policy frameworks.
\end{enumerate}
To ensure consistency, both authors coded the same 70 articles, finding 82\% agreement in inclusion decisions. We then discussed and refined the criteria before continuing. In total, 57 articles were included from $S_0$.

\paragraph{Snowballed set ($S_1$)} To address blindspots in our keywords and corpora, we gather additional articles referenced within $S_0$ articles, regardless of where they were published ($n=72$), and then reapply the inclusion criteria to ensure relevance. 
This results in 38 additional articles from $S_1$.

We further narrow our scope with two categorisations on the 95 articles from $S_0 + S_1$:
\paragraph{Dominant contribution types} We categorise articles into: (i) those that \textit{evaluate} or benchmark model's capabilities, ethics or worldviews ($n=14$); (ii) those that \textit{predict} human preferences and values from social media data using specialised models not intended for other downstream generative or classification tasks ($n=9$); and (iii) those that \textit{train} or seek to align models with a general notion of human preferred or valued text ($n=72$). The last category is the focus of our review.\footnote{There is overlap between categories---papers that fine-tune LLMs according to human preferences also evaluate these trained models. See \cref{sec:other_conts} for further detail.}

\paragraph{Use of LLMs} For the purposes of our review, we define LLMs as any encoder-only, decoder-only or encoder-decoder model that is pre-trained with self-supervised learning over large internet corpora. As a rule of thumb, we consider BERT \cite{devlinBERT2019} and ELMO \cite{peters2018deep} among the first LLMs; so, any articles published before 2018 fall outside our definition. Of the 72 \textit{train} articles, we cover 22 articles published in the pre-LLM era in our review of The Past (\cref{sec:the_past}) and 50 articles using LLMs in The Present (\cref{sec:the_present}).

\subsection{Coding Articles}
We examine each article under two main themes.\footnote{The full code book is presented in \cref{sec:code_book}.} The \textbf{Conceptual} theme documents the motivations for collecting human feedback; the definitions of human preferences or values; whether these are understood as universal or contextual/cultural; and the level of interpretative freedom in applying preference or value concepts. The \textbf{Methodological} theme covers sub-categories on (i) annotation or labour force details, such as how feedback was collected and from whom; and (ii) modelling details, such as how feedback is integrated into training and evaluation phases, and the target task. We also collect procedural details on academia vs industry authorship, whether empirical artefacts (data and/or models) are publicly available, and if (to the best of our knowledge) the paper has been peer-reviewed.

\section{The Past}
\label{sec:the_past}
In this section, we review 22 articles released between 2014-2019 that use human feedback but with older generation model architectures. Highlighting these works ensures that foundational research is adequately attributed for advancements in today's models, and demonstrates the evolution from indirect or proxied human feedback.

\subsection{Conceptual Classification}
 None of the articles released in this period seek alignment to human values. Instead, they generate text according to human preferences in machine translation \cite{mirkinMotivating2015, mirkinPersonalized2015, lawrenceCounterfactual2017, nguyenReinforcement2017, rabinovichPersonalized2017a, kreutzerReliability2018} and dialogue \cite{liDeep2016, moPersonalizing2016, liLearning2017, wangPredicting2017, liuDialogue2018, jaquesWay2019}. 
Preferences are defined in both personal and universal contexts, reflecting the persistent difficulties of separating the two.
\citet{ficlerControlling2017} focus on modulating formality depending on context, while others focus on the personalisation of language models, such as reflecting author personality in machine translation \cite{mirkinMotivating2015, mirkinPersonalized2015, rabinovichPersonalized2017a}; providing financial recommendations via chat bots \cite{hengstReinforcementLearningPersonalized2019}; or enabling customised online shopping \cite{moPersonalizing2016}. Most studies target human preferences assumed to be commonly-held and stable, such as word order \cite{futrellRNNs2019}, sense making \cite{dedeynePredicting2016, seminckComputational2017} and vocabulary matching \cite{campanoComparative2014, dubuissonduplessisAutomatic2017}. In contrast, \citet{nguyenReinforcement2017} and \citet{kreutzerBandit2017} acknowledge the noisiness of human feedback but attempt to extract a single, unified preference.

\subsection{Methodological Classification}
Most articles use pre-transformer recurrent neural networks such as LSTMs \cite{hochreiterLongShorttermMemory1997, vaswaniAttentionAllYou2017}. Few articles use direct human feedback, mostly in information retrieval tasks. In two cases, humans answer a series of yes/no questions to provide a more expressive reward for reinforcement learning (RL) \cite{liDialogue2017, lawrenceImproving2018}. \citet{dhingraEndtoEnd2017} use requests for additional information to form better queries with a binary `success/failure' reward. \citet{lawrenceCounterfactual2017} and \citet{lawrenceImproving2018} compare forms of human feedback, finding cardinal feedback to be more useful than pairwise comparison. 

Human feedback is an expensive and time-consuming source of data to collect, which motivates efforts to find reliable proxies \cite{lawrenceCounterfactual2017, nguyenReinforcement2017}. Implicit feedback methods attempt to utilise naturally-occurring signals in human interactions, such as sentiment \cite{wangPredicting2017, jaquesWay2019} and response length \cite{campanoComparative2014}. Other articles define rules on desirable dialogue properties, such as length \cite{liDeep2016}, 
vocabulary alignment \cite{ dubuissonduplessisAutomatic2017}, or tone \cite{ficlerControlling2017}, and score the agent for achieving them. Only \citet{liDeep2016} apply RL to further train the model from the feedback. 

Simulating human feedback is also a commonly-used and cost effective approach where `feedback' is generated by measuring similarity to the gold standard in pre-existing, human-labelled datasets. Parallel translation corpora are a common source of gold demonstrations, e.g., translated TED talks \cite{mirkinMotivating2015, mirkinPersonalized2015, nguyenReinforcement2017} 
or European Parliament speeches \cite{kreutzerBandit2017, rabinovichPersonalized2017a}. User simulators typically use a `success/failure' score for RL \cite{moPersonalizing2016, liuDialogue2018}, while `gold standard' approaches use a more complex loss function on output similarity \cite{mirkinPersonalized2015, nguyenReinforcement2017}.






\section{The Present}
\label{sec:the_present}
Turning our attention to the heart of our review, this section discusses the 50 papers that incorporate human feedback to steer LLM behaviours.
\subsection{Conceptual Classification}
We first seek to understand \textbf{why human feedback is collected}. The motivations for eliciting feedback form two groups. The first group generally seeks \textit{value alignment}, i.e., some notion of steering language models towards producing societally-desirable text \cite{zhaoEthicalAdvice2021, liuDExperts2021}. We note a variety of vague goals such as to reduce ``non-normative'' \cite{pengReducing2020} or ``immoral''  text \cite{liuSecond2023}; to generate more ``pro-social'' \cite{liuAligning2022} or ``legitimate'' text \cite{bakkerFinetuning2022}; or to encourage that LLM technologies have a ``positive impact on society'' \cite{liuLanguages2023}. Specific motivations include minimising toxic or offensive language \cite{dinanBuild2019, xuBotAdversarial2021, juLearning2022, scheurerTraining2022, korbakPretraining2023}; improving safety \cite{liuDExperts2021, xuRecipes2021, thoppilanLaMDA2022, ganguliRed2022, jinWhen2022}; adapting to ethical or moral scenarios \cite{forbesSocial2020, jiangCan2022, jinWhen2022}; or achieving political ideological balance \cite{liuMitigating2021}. The broad definitions of value alignment mostly assume some universality of value dimensions.\footnote{An example of a vague definition is that a ``value-aligned system should make decisions that align with human decisions in similar situations and, in theory, make decisions which are unlikely to be harmful'' \cite[][p.~1]{nahianLearningNormsStories2020}.} However, some do seek to align LLMs to specific groups, sets of values or according to cultural context \cite{solaimanProcess2021, qiuValueNet2021, bangEnabling2022}.

The second group of articles is motivated by more practical target concepts of improving model capabilities, particularly when clear optimisation metrics or programmatic rewards are lacking \cite{zieglerFineTuning2019, wuRecursively2021, glaeseImproving2022, baiConstitutional2022}. Motivations often revolve around generating high-quality or human-preferred outputs \cite{gaoAPRIL2018, bohmBetter2019, jaquesHumancentric2020, stiennonLearning2020, wangNonParametric2021, scheurerTraining2022, nguyenMake2022, xuLearning2022}, without much discussion of why this matters or whether humans agree amongst themselves what is ``high-quality''. Specific target attributes include minimising repetitiveness \cite{aroraDirector2022}; increasing coherence \cite{luBoosting2022}, usefulness \cite{liuDExperts2021}, engagingness \cite{gaoDialogue2020, xuRecipes2021, luBoosting2022}, or interest \cite{thoppilanLaMDA2022}; and producing human-like conversations \cite{hancockLearning2019, jaquesHumancentric2020}. Some seek greater explainability and factuality in generated text \cite{nakanoWebGPT2021, menickTeaching2022, scheurerTraining2022, thoppilanLaMDA2022} or correctness in code \cite{korbakPretraining2023}. Preferences can also be elicited for customisation and personalisation \cite{majumderGenerating2019, zhouNaRLE2021, dengPersonalized2021}.

The boundary between preference- and value-motivated aims is not always clear-cut. Commonly-adopted mixed motivations include helpful, honest and harmless behaviours, introduced by \citet{askellGeneral2021} and adopted by others \cite{baiConstitutional2022, baiTraining2022, bakkerFinetuning2022, menickTeaching2022}. \citet{thoppilanLaMDA2022} target safety, quality and groundedness---concepts that similarly blur the lines between practical preferences and value-laden judgements. Even for instruction-tuning articles motivated by inferring user intent, what \citet{ouyangTraining2022} call ``explicit'' and ``implicit'' intent is synonymous with the helpfulness versus honesty/harmlessness distinction.\footnote{\citet[][p.2]{ouyangTraining2022} include ``explicit intentions such as following instructions and implicit intentions such as staying truthful, and not being biased, toxic, or otherwise harmful.''}

\subsection{Methodological Classification}
We primarily discuss how feedback is collected (\cref{sec:collect_feedback}) and integrated into LLMs (\cref{sec:integrating_feedback}). We additionally present an overview of target tasks and evaluation methods in \cref{sec:additional_appendix_info}. 
\subsubsection{Collecting Feedback}
\label{sec:collect_feedback}
First, we address \textbf{how feedback is collected}.
Explicit comparisons collected on model outputs are used to reveal the preferences of human raters \cite{gaoAPRIL2018, zieglerFineTuning2019, askellGeneral2021, jaquesHumancentric2020, stiennonLearning2020, ganguliRed2022, glaeseImproving2022}.\footnote{Usually, ratings are collected between two outputs \cite{baiConstitutional2022, ganguliRed2022} but others use four \cite{zieglerFineTuning2019} or even up to 9 items for comparison \cite{ouyangTraining2022}. A null vote is predominately not included (\textit{neither of these outputs are good}) which may be a particular problem for harm assessment \cite{ganguliRed2022}---though some address ties in preference strength \cite[e.g.,][]{baiTraining2022, menickTeaching2022}.} More fine-grained feedback includes binary or Likert scale questions on text attributes \cite{nakanoWebGPT2021, menickTeaching2022, thoppilanLaMDA2022}; natural language comments \cite{juLearning2022, scheurerTraining2022}; or edits \cite{hancockLearning2019, luBoosting2022, liuSecond2023}. Ideal demonstrations are used to ground norm-dependent or ethical judgements \cite{forbesSocial2020, zhaoEthicalAdvice2021, pyatkinReinforced2022, jiangCan2022, jinWhen2022}, or in combination with ratings to prime model behaviour \cite{nakanoWebGPT2021, wuRecursively2021, ouyangTraining2022, bakkerFinetuning2022}. Several articles collect negative feedback via adversarial examples \cite{dinanBuild2019, xuBotAdversarial2021, xuRecipes2021, glaeseImproving2022}. \citet{xuLearning2022} test various feedback types including binary, free-form conversation, and fine-grained failure modes.

Human input can be further removed from directly assessing model outputs. For example, simulating feedback with an ``oracle'' assumed to prefer specific text attributes measured via automated metrics \cite{wangNonParametric2021, nguyenMake2022, korbakPretraining2023} or predictions from a separate classifier \cite{pengReducing2020, liuMitigating2021}. In \citet{baiConstitutional2022} human input defines the constitution but AI feedback is used to implement it during training. A seed of human generated examples guiding synthetic data generation also applies elsewhere \cite{bangEnabling2022, castricatoRobust2022, honovichUnnatural2022, wangSelfInstruct2022}. Other articles adopt human labels on pre-existing datasets \cite{bohmBetter2019, liuDExperts2021, aroraDirector2022, jiangCan2022},
or leverage implicit feedback data from stories \cite{nahianLearningNormsStories2020} and social media such as Reddit or StackOverflow \cite{gaoDialogue2020, askellGeneral2021, baiTraining2022}. Feedback can also be inferred from certain language patterns or emotive attributes in conversations with human partners 
\cite{hancockLearning2019, zhouNaRLE2021}.

Second, we address \textbf{who feedback is collected from}. Almost all articles use crowdworkers for training and/or evaluation, recruited from a variety of sources---including MTurk \cite{nahianLearningNormsStories2020, pengReducing2020, jaquesHumancentric2020, liuDExperts2021, liuMitigating2021, qiuValueNet2021, baiTraining2022, ganguliRed2022, jinWhen2022, xuLearning2022, juLearning2022}; Upwork \cite{stiennonLearning2020, baiTraining2022, ganguliRed2022}; ScaleAI \cite{ouyangTraining2022, stiennonLearning2020, zieglerFineTuning2019}; and SurgeAI \cite{solaimanProcess2021, nakanoWebGPT2021, baiConstitutional2022}. With `in-the-wild' social media data, social media users unknowingly become the `raters' \cite{gaoDialogue2020, askellGeneral2021, baiTraining2022}. \citet{ouyangTraining2022} include OpenAI API users as ``demonstrators''. At least 13 articles rely on their authors for a variety of tasks:\footnote{In \citet{scheurerTraining2022}, the article relies on two authors to provide the feedback data and two other authors to do the human evaluation experiments.}
writing seeds to scale synthetic data \cite{honovichUnnatural2022, wangSelfInstruct2022}; hand-crafting conditioning prompts \cite{askellGeneral2021,  glaeseImproving2022}; defining a constitution \cite{baiConstitutional2022}; specifying topics or starter questions \cite{solaimanProcess2021, bakkerFinetuning2022}, and ethical scenarios \cite{zhaoEthicalAdvice2021}; conducting evaluation \cite{stiennonLearning2020, ganguliRed2022, luBoosting2022} or generating benchmarks \cite{baiTraining2022}; and compiling training tasks for crowdworkers \cite{qiuValueNet2021, glaeseImproving2022}. Even without direct involvement, authors can influence feedback collection by writing annotator guidelines.

\subsubsection{Integrating Feedback}
\label{sec:integrating_feedback}
\paragraph{RL with Direct Human Feedback}
A reward signal can first be extracted by asking \textit{actual humans} about their preferences for model outputs then embedded into the LLM via RL fine-tuning. 
The general recipe is as follows: \textbf{(Step 1)}: Either take an ``off-the-shelf'' pre-trained LLM \cite{luBoosting2022}; Or adapt this model via prompt-guiding \cite{askellGeneral2021, bakkerFinetuning2022,  baiTraining2022, glaeseImproving2022} or supervised fine-tuning and behavioural cloning over ideal demonstrations \cite{zieglerFineTuning2019, stiennonLearning2020, nakanoWebGPT2021, ouyangTraining2022, menickTeaching2022}.\footnote{For example, \citet{luBoosting2022} use a SOTA chinese chatbot; \citet{nakanoWebGPT2021} start with GPT-3 architecture (760M, 13B, 175B); \citet{baiTraining2022} use the Context-Distilled LM (52B) from \citet{askellGeneral2021}; \citet{glaeseImproving2022} hand-author prompts to demonstrate `good' behaviour in a Dialogue Prompted Chincilla model (70B); \citet{stiennonLearning2020} start with versions of GPT-3 (1.3B, 6.7B) fine-tuned on filtered TL;DR Reddit dataset;  \citet{zieglerFineTuning2019} use a fine-tuned GPT-2 model; \citet{menickTeaching2022} use supervised fine-tuning on Gopher family models with examples rated positively by labellers; and \citet{ouyangTraining2022} fine-tune GPT-3 on demonstrations of desired behaviours.}
 \textbf{(Step 2)}: Generate multiple outputs from this model, and employ crowdworkers to create a comparisons dataset. \textbf{(Step 3)}: Train a preference reward model (PM) on this feedback so ``better'' items are assigned higher score \cite{baiTraining2022}---either a scalar reward for a given item or an ELO score i.e., the log odds that A $\succ$ B \cite{stiennonLearning2020, nakanoWebGPT2021, glaeseImproving2022}. The PM can be pre-trained on naturally-occurring text and ratings e.g., from Reddit or Stackoverflow \cite{askellGeneral2021, baiTraining2022}. \textbf{(Step 4)}: Fine-tune a RL policy (another LM) that generates text autoregressively, whilst the PM provides a reward signal. Often, the policy is updated using the PPO algorithm \cite{zieglerFineTuning2019, stiennonLearning2020, nakanoWebGPT2021} and a KL-penalty term is applied to control deviations from the base model \cite{jaquesWay2019, zieglerFineTuning2019, stiennonLearning2020, nakanoWebGPT2021, menickTeaching2022, ouyangTraining2022, liuSecond2023}. This pipeline can be implemented in offline, online or batched settings \cite[see][]{zieglerFineTuning2019}.
Modifications to the recipe include using recursive subtasks \cite{wuRecursively2021}; applying a rule reward model in addition to the PM to penalise undesired outputs \cite{glaeseImproving2022}; or using the PM to re-rank or reject sample generations from the supervised model \cite{askellGeneral2021, nakanoWebGPT2021, glaeseImproving2022, ganguliRed2022, baiTraining2022, xuLearning2022, bakkerFinetuning2022}, which can match or outperform optimising a model via RL \cite{menickTeaching2022, thoppilanLaMDA2022}.

\paragraph{RL with Indirect Human Feedback}
A reward can be \textit{inferred} without directly asking humans about their preferences over model outputs. These articles skip the step of training a PM from comparisons data, and instead infer preferences from textual attributes of outputs \cite{jaquesHumancentric2020, zhouNaRLE2021}. It varies how far removed the human input is, for example in designing the constitution \cite{baiTraining2022}, in determining the automated metric \cite{nguyenMake2022, korbakPretraining2023} or in compiling the word lists to measure political bias \cite{liuMitigating2021}. Often another model is treated as the `oracle' to simulate human rewards---\citet{gaoAPRIL2018}, for example, simulate preferences on two summaries with perfect, noisy and logistic noisy ``oracles'' based on ROGUE scores; \citet{wangNonParametric2021} take the reward as human revisions from parallel machine translation corpora; while others deploy the rewards from a value, moral or toxicity classifier trained on crowdworker labels to reinforce a generator \cite{qiuValueNet2021, liuAligning2022, castricatoRobust2022, pyatkinReinforced2022}.

\paragraph{Generator and Discriminator}
Some use a unified generator and classifier step to steer the LLM away from undesirable text \cite{aroraDirector2022}, for example using other fine-tuned LLMs to modify the predicted probability in a base model for the next token at decoding time \cite{liuDExperts2021}. A combined model that functions as a generator and a discriminator can be trained sequentially \cite{thoppilanLaMDA2022} or jointly \cite{luBoosting2022}.

\paragraph{Preference Pre-training} \citet{korbakPretraining2023} argue that incorporating human feedback in supervised or RL fine-tuning phases is suboptimal. Instead, they approach alignment in the pre-training phase of GPT-2, finding that conditional training is the most effective pre-training objective, and is more robust than later fine-tuning an already pre-trained model.

\paragraph{Preference Fine-Tuning}
Human feedback can be incorporated via supervised fine-tuning \cite{ hancockLearning2019, nahianLearningNormsStories2020, jiangCan2022}. For example, \citet{gaoDialogue2020} apply contrastive learning with a GPT-2 based dialogue model over 133M pairs of human feedback data with a loss designed to simultaneously maximise the positive sample score and minimise the negative score. \citet{liuLanguages2023} use ``chain of hindsight'' fine-tuning to include both positive and negative feedback. Fine-tuning data is often filtered relative to the value or preference goal \cite{solaimanProcess2021, xuLearning2022, bangEnabling2022}. \citet{pengReducing2020} instead train a reward model (normative text classifier) but this reward is applied to the loss and backpropagated during fine-tuning.

\paragraph{Prompting}
Prompting is a simple way to align LLMs with specified human preferences and values. \citet{jinWhen2022} cast moral situations as multi-step prompts to elicit chain of thought reasoning in InstructGPT, while \citet{zhaoEthicalAdvice2021} use zero- and few-shot prompts for responsive questioning on unethical behaviours. \citet{askellGeneral2021} show that using a long prompt (4,600 words) from ideal author-written conversations is an effective alternative to fine-tuning in data-constrained scenarios. They also use context distillation by training a new LLM to replicate the behaviour of another LLM that is using a specific prompt.
\section{Challenges and Recommendations for the Future}
Drawing on our analysis of the reviewed papers, we identify five key challenges for future researchers. These challenges are divided into conceptual and practical issues. The conceptual challenges (C1-C3) revolve around the difficulty of specifying a clear set (or sets) of preferences and values. Even assuming the resolution of the conceptual challenges, practical challenges remain in converting conceptual ideals into empirical signals, which in turn steer language model behaviours.

\paragraph{(C1) Preferences and values are not universal}
`Aligning' a language model requires a set of desired preferences or values to align with; but specifying such a set is an unresolved problem. 
One popular approach is to specify a minimal set of ostensibly unobjectionable and widely-shared values, such as helpfulness, honesty and harmlessness \cite{baiTraining2022, baiConstitutional2022, thoppilanLaMDA2022}. However, these values are only unobjectionable because they are abstract and not precisely defined \cite{kirkEmpty2023b}. These terms can be considered what Levi-Strauss and Laclau call `empty signifiers' \cite{levi-straussIntroduction1987, laclauPopulist2005}; terms that are viewed positively but are inscribed with different meanings by different people.
For example, when \citet{baiTraining2022} design a constitution to produce outputs as ``ethical and harmless as possible'', this can have varying interpretations based on an individual's own ethical frameworks and sociocultural background. Establishing priorities over sets of preferences or values to embed in LLMs, and ensuring consistent interpretation of conceptual meaning across people, is a persistent challenge which cannot alone be resolved via purely technical solutions. One possible approach is to draw on legal theory, and values protected in human rights law \cite{solaimanProcess2021}. Translating abstract shared values into decisions is a core function of legal systems and legal theory offers a long history of scholarship which combines the philosophical and practical. 
One approach along these lines was proposed by \citet{kirk2023personalisation} which applies a principle of subsidiarity to govern the personalisation of generative AI systems for different use cases. 
We also advocate for anchoring closely to existing legal systems as a matter of democratic principle: it is dangerous for moral and value judgements with broad societal impacts to be made by small independent groups.

\paragraph{(C2) Preferences and values are inconsistently defined} Although the terminology of `preferences' and `values' implies some difference between the two, the conceptual basis and normative implications of this distinction is often unclear.
Colloquially, values are understood to be stronger than preferences, and potentially carry greater normative weight as guiding principles or life goals \cite{fischerPersonalityValuesCulture2017}. As such, users may have greater concerns about an LLM misaligned with their values than with their preferences; So, it is important to be clear about which is being discussed. 
Within the broad terms, there are many meanings: `preferences' have been defined as `instrumental utility' \cite{dubuissonduplessisAutomatic2017, gaoAPRIL2018, nguyenMake2022}, `stylistic taste' \cite{mirkinPersonalized2015, seminckComputational2017, jaquesHumancentric2020}, and `behavioural principles' \cite{baiConstitutional2022, castricatoRobust2022}. `Values' definitions are based on `instrumental and intrinsic value' \cite{askellGeneral2021}, `acceptable social behaviours' \cite{forbesSocial2020, bangEnabling2022}, or `making decisions which are unlikely to be harmful'
\cite{nahianLearningNormsStories2020}. The differences between individual (subjective) and global (objective) preferences is often blurred---for example, which properties of a ``better'' summary are universal, and which depend on subjective appreciation, like writing style and tone. Clearer definitions of preferences and values in the context of alignment would serve to motivate and clarify \textit{what} we are aligning LLMs to.

\paragraph{(C3) Human feedback is inherently incomplete}
Alignment via human feedback ultimately relies on LLMs being capable of successfully generalising from few examples to new cases and domains. This is because the space of possible behaviours over which to collect feedback is prohibitively large and is not fully known. An open question is the extent to which models generalise from partial human feedback, especially when presented with data that is completely out-of-domain for their training or at the margins of its distribution. 
For instance, if an LLM is trained with examples of safe responses to user prompts which deny the Holocaust, it may generalise to different expressions of the same canonical request. However, it will not necessarily learn how to handle denials that the earth is round and denials of vaccine efficacy, or have domain expertise for other harmful requests, such as users who ask how to make a bomb or bio-weapon. Human values are considered to be fairly stable guiding principles that manifest similarly across situations for a given individual \cite{fischerPersonalityValuesCulture2017} but the same generalisation cannot be guaranteed of LLMs. 

Several related epistemological issues arise from technical details of the methods being used. Reinforcement learning 
introduces a path-dependence problem, where the particular order in which feedback is given may change the quality of the final results. As a result, it is difficult to know whether a local optimum is reached which is notably worse than the global optimum. 
With any form of learning from feedback, language models may also overfit or appear to be aligned externally, but have persistent internal misalignment which manifests subtly in cases more distant from the training data \cite{perezDiscovering2022}. These challenges become yet more convoluted when dealing with more complex tasks---an issue that \citet{bowmanMeasuring2022} examine in their discussion of scalable oversight. 

\paragraph{(C4) Operationalising a ``good'' output is difficult} Even if a shared set of values could be agreed upon, converting these thick normative concepts into signals that models can use, such as by collecting annotator ratings, is hard. Complex goal operationalisation is itself a motivator for collecting feedback---when humans may not be able to articulate their preferences or write ideal demonstrations but can rate outputs, a kind of ``I know it when I see it'' logic. However, training with human feedback involves moving values and preferences from the abstract to particular survey or rating instruments, reinforcing differences in interpretation.
To reduce disagreements, some authors write very prescriptive and/or comprehensive guidelines for the task in order to ``make comparisons as unambiguous as possible'' \cite[][p.18]{nakanoWebGPT2021}. Several papers still find low inter-annotator agreement with such prescriptive approaches \cite{stiennonLearning2020, glaeseImproving2022, baiTraining2022, ouyangTraining2022}. In other cases, annotators are explicitly allowed to use their own subjective assessment, to ``interpret these concepts as they see fit'' \cite[][p.4]{baiTraining2022}, but then agreement between annotators is no longer ensured. When multiple text attributes affect annotators' preferences, it is hard to pin down what we are actually measuring. For example, \citet{stiennonLearning2020} and \citet{wuRecursively2021} condition their evaluation question as ``how good is this summary, given that it is X words long?''. Hypothetically, if ``good'' is subjective then the question should be ``how good is this summary for individual Y?''. Some guidelines do ask annotators to role-play or put themselves in the shoes of others, for example to infer the intent of a prompt \cite{ouyangTraining2022} or question \cite{nakanoWebGPT2021}, but this may introduce further problems, especially for value-laden judgements where the rater may have a biased interpretation of how to apply another person's values \cite{qiuValueNet2021}.

To aid transparent communication, it should be clearly documented whether researchers aspire to follow the prescriptive or subjective paradigm of data annotation, rather than leaving it unspecified \cite{rottger2022two, kirkEmpty2023b}. Increased interdisciplinary communication with practitioners in other fields would impart wisdom on measuring the perspectives and behaviours of human subjects. For example, Human-Computer Interaction literature shows how interfaces and incentives can be optimally designed to avoid participant response bias \cite{deng2010affect, dell2012yours, hsieh2016you}; Experimental psychology and behavioural economics research show how the presentation of scales and order effects influence ratings \cite{friedman1994biasing, maeda2015response, westland2022information} and that human preferences are unstable, intransitive and vulnerable to experimental artefacts \cite{tversky1969intransitivity,lacy2001theory, chiesa2008making, lee2009search, chuang2015stability}. Researchers should consider techniques to model the noise and distribution of human feedback \cite{juLearning2022} or establish post-hoc consensus \cite{bakkerFinetuning2022}, rather than ignoring disagreement by aggregating responses. However, there are trade-offs: the specific nuances and minutiae captured in fine-grained feedback might heighten biases and reduce generalisability when drawn from unrepresentative samples---which we now discuss.

\paragraph{(C5) Crowdworkers and social media users are neither representative nor sufficient}
A degree of subjectivity persists even with prescriptive guidelines and well-designed experimental instruments; So, outcomes critically depend on who is interpreting value or preference-based concepts. In the majority of articles, fewer than 100 humans are employed to guide or evaluate language model behaviours \cite{jaquesHumancentric2020, stiennonLearning2020, nakanoWebGPT2021, menickTeaching2022, baiTraining2022, ouyangTraining2022, jinWhen2022, pyatkinReinforced2022}, which is concerning for ethically or morally ambiguous scenarios. It is striking that so few voices have so much power in shaping LLM behaviours---in \citet{baiTraining2022} just 20 humans contributed 80\% of the feedback data, and in \citet{nakanoWebGPT2021} the top 5 humans contributed 50\%. Workforces employed for evaluation are similarly small, with some employing <25 workers \cite{scheurerTraining2022, castricatoRobust2022, gaoAPRIL2018, liuLanguages2023}.
Overwhelmingly, these humans are US-based, English-speaking crowdworkers with Master's degrees and between the ages of 25-34. This results in a non-democratic and non-diverse feedback process, termed ``the tyranny of crowdworker'' by \citet{kirk2023personalisation}, which has been shown to introduce political and religious biases in model behaviours \cite{perezDiscovering2022}. The limitations of relying on the subjective interpretations of a small and non-representative work force are exacerbated by inadequate documentation. Only nine out of 50 papers provided solid documentation, such as demographic breakdowns \cite{stiennonLearning2020, thoppilanLaMDA2022, baiTraining2022, ganguliRed2022, glaeseImproving2022, jinWhen2022, liuAligning2022, ouyangTraining2022, liuSecond2023}. Others provide high-level details of the rater pool such as number of workers, hiring platform, or aggregate demographics. The majority of articles do not document their workforce, nor discuss sample biases or annotator artefacts.

When soliciting human feedback, attempts should be made to  diversify who is given a voice, such as by applying democratic or jury-based principles in how these voices are weighted \cite{gordon2022jury} and by employing bottom-up participatory approaches \cite{martinjrParticipatoryProblemFormulation2020, birhanePower2022, zytkoParticipatory2022, derczynskiAssessing2023}; Or to seek top-down sampling that better represents the population being studied 
\cite{bakkerFinetuning2022}. Mirroring the move in other areas of NLP to document and explore annotator disagreement \cite{aroyo2015truth, geva2019we, nie2020can, prabhakaran2021releasing, davani2022dealing}, each item of feedback should be associated with a pseudo-anonymised annotator ID. So far as privacy allows, documentation of annotator background should be provided in a data statement \cite{bender2018data}.

\section{Conclusion}
This review provided an overview of incorporating human feedback into LLMs, with a focus on subjective preferences and values that lack `ground truth alignment'. 
We have witnessed two notable shifts in the field from past to present---first, a move away from specialist systems towards general purpose language agents capable of handling many NLP subtasks via instruction or open-ended dialogue; second, more use of direct human feedback which surpasses the limitations of user simulations or automated metrics. 

While the shift to incorporate human voices directly into LLM development is welcome, it introduces new challenges that require careful navigation. Some challenges are more tractable than others---for example, practitioners will always have to deal with the complexities and intricacies of unstable and idiosyncratic preferences across end users of their model, but can take practical steps to better approximate this distribution by diversifying the recruitment of feedback providers.

External scrutiny is crucial to ensure the integrity and reliability of research efforts. Our review shows that many influential papers lack open and externally-validated peer review, especially those published by big industry labs like Google DeepMind, Anthropic, Google, and OpenAI. Furthermore, the majority of reviewed papers do not release model artefacts, or only do so behind a paywalled API. To foster progress, we advocate for a greater degree of open, interdisciplinary and democratically-grounded discussion on how humans can meaningfully shape future LLM behaviours in a way which is well-bounded, operationalisable, and equitable.
\section{Limitations}
\label{sec:limitations}
We discuss limitations associated with our review:

\paragraph{Applying topic exclusion} We exclude articles on the basis of being unrelated to the topic of value or preference alignment, but found it consistently difficult to draw a clear distinction between articles in and out of scope. For validation purposes, we had both reviewers read a portion of the articles, and found the cases of disagreement helpful to highlight this challenge. One such example was with two articles using similar methods to approach translation which we initially classified differently,  \citet{kreutzerBandit2017} and  \citet{lawrenceCounterfactual2017}. The papers primarily treat translation as an objective task focused on BLEU scores, which would make them out of scope. However, translation inherently involves stylistic and subjective judgements, and the methods developed seek to replicate these judgements from the training data, blurring the distinction. Honesty is another target concept with these issues---whether in-text claims are referenced is fairly black and white, but whether an end user ascribes more trust to the system because of these references is subject to idiosyncratic epistemology. We use this case to highlight the weaknesses of creating a dichotomy between subjective and objective preferences in practice.

\paragraph{Related NLP subfields} A related issue is where to draw the line for what is and is not in scope. Some narrowing was needed to make the review focused, feasible and instrumentally useful to future practitioners. However, technically fairness and bias are values of intrinsic utility---hence their inclusion in many AI principles around the world \cite{jobin2019global}. That said, there exists a very wide and distinct literature on fairness and bias in LLMs that would be too expansive for this review \cite[see, e.g.,][]{chang2019bias, lucyGenderRepresentationBias2021, abidPersistentAntiMuslimBias2021, nadeemStereoSetMeasuringStereotypical2021, kirkBiasOutoftheBoxEmpirical2021, smithSorryHearThat2022}. Similarly, there are sub-literatures on other aspects of LLM behaviours---such as toxicity \cite{gehmanRealToxicityPromptsEvaluatingNeural2020a, welblChallengesDetoxifyingLanguage2021}, truthfulness \cite{linTruthfulQAMeasuringHow2022} or hallucination \cite{jiSurveyHallucinationNatural2022}. We explicitly focus on papers that target some notion of human preferences and values in their motivations, but the challenges raised from our review can be applied to other fields which similarly suffer from subjectivity in interpretative scope---e.g., carefully deciding who the human labellers are and what guidelines govern their interpretation of concepts.

\paragraph{Blindspots in reviewed articles} Blindspots come from a number of sources. First, \textit{keyword and corpora blindspots}: We ground our initial review on articles from arXiv and ACL using a set of defined keywords. We attempt to  mitigate blindspots by snowballing related and relevant articles outside our initial collection; however, it is almost certain that we have missed some papers in the field as a whole. Second, \textit{language blindspots}: Our review only contains articles written in English, limited by the expertise of authors who acted as the coders. This bias however reflects the dominance of English in academic publishing in general, but English language proficiency may gatekeep the concepts and voices already contributing to LLM development. Third, \textit{community blindspots}: we only look at academic papers---but issues surrounding large language model behaviours or alignment have become a hot topic of conversation on blogs and social media forums. We inherently exclude such discussions from these other stakeholder communities. Fourth, \textit{modality blindspots}: there is a rich history of using RL to align models in other modalities, such as delegate agents acting in toy or game worlds \cite[see, e.g.,][]{christiano2017deep}. We do not cover the insights from these related literatures. Finally, \textit{temporal blindspots}: research into LLMs is a fast-paced field---in one week, there can be as many as 500 articles posted on the cs.CL subclass of arXiv. Inevitably, other influential articles have been released after the review was completed and more were released during its peer review period. A good example of this is \citet{rafailovDirect2023} who introduce Direct Preference Optimisation, a technique that could substantially change how people approach feedback learning in the future. Other relevant papers that appeared after the cut-off for this review include \citet{dongRAFT2023, hoskingHuman2023, liuTraining2023, liuChain2023, songPreference2023, yuanRRHF2023, wuFineGrained2023, zhouLIMA2023}. With the field's rapid developments, any review paper runs the risk of lagging behind the latest research. However, given the substantial number of articles that we did review, we expect many of the general findings and highlighted challenges to apply in upcoming future work.

\paragraph{External scrutiny of reviewed articles} We consciously made the decision to include articles which have not yet been peer reviewed to stay ahead of the curve with early-released pre-prints and also to track industry contributions (which are often not externally peer reviewed). In the 22 papers appearing the The Past section, 18 were peer reviewed. Of the 50 papers appearing in The Present section, only 21 were clearly peer-reviewed. It is a contentious issue that many influential papers lack standard practices of external scrutiny and rigorous academic backstops, though often industry-authored papers do undergo a process of internal review before a preprint is released.

\section*{Acknowledgements}
This paper received funding from a MetaAI Dynabench grant as part of a research agenda on optimising feedback between human-and-model-in-the-loop. H.R.K.’s PhD is supported by the Economic and Social Research Council grant ES/P000649/1. A.M.B.'s PhD is supported by the Clarendon Fund. P.R. received funding through the INDOMITA project (CUP number J43C22000990001) and the European Research Council (ERC) under the European Union’s Horizon 2020 research and innovation program (No. 949944, INTEGRATOR).

\bibliography{out_review, in_review}

\begin{thebibliography}{148}
\expandafter\ifx\csname natexlab\endcsname\relax\def\natexlab#1{#1}\fi

\bibitem[{Abid et~al.(2021)Abid, Farooqi, and
  Zou}]{abidPersistentAntiMuslimBias2021}
Abubakar Abid, Maheen Farooqi, and James Zou. 2021.
\newblock \href {https://doi.org/10.1145/3461702.3462624} {Persistent
  {{Anti-Muslim Bias}} in {{Large Language Models}}}.
\newblock In \emph{Proceedings of the 2021 {{AAAI}}/{{ACM Conference}} on
  {{AI}}, {{Ethics}}, and {{Society}}}, {{AIES}} '21, pages 298--306, {New
  York, NY, USA}. {Association for Computing Machinery}.

\bibitem[{Arora et~al.(2022)Arora, Shuster, Sukhbaatar, and
  Weston}]{aroraDirector2022}
Kushal Arora, Kurt Shuster, Sainbayar Sukhbaatar, and Jason Weston. 2022.
\newblock \href {https://aclanthology.org/2022.aacl-main.39} {Director:
  {Generator}-{Classifiers} {For} {Supervised} {Language} {Modeling}}.
\newblock In \emph{Proceedings of the 2nd {Conference} of the {Asia}-{Pacific}
  {Chapter} of the {Association} for {Computational} {Linguistics} and the 12th
  {International} {Joint} {Conference} on {Natural} {Language} {Processing}
  ({Volume} 1: {Long} {Papers})}, pages 512--526, Online only. Association for
  Computational Linguistics.

\bibitem[{Aroyo and Welty(2015)}]{aroyo2015truth}
Lora Aroyo and Chris Welty. 2015.
\newblock Truth is a lie: Crowd truth and the seven myths of human annotation.
\newblock \emph{AI Magazine}, 36(1):15--24.

\bibitem[{Askell et~al.(2021)Askell, Bai, Chen, Drain, Ganguli, Henighan,
  Jones, Joseph, Mann, DasSarma, Elhage, Hatfield-Dodds, Hernandez, Kernion,
  Ndousse, Olsson, Amodei, Brown, Clark, McCandlish, Olah, and
  Kaplan}]{askellGeneral2021}
Amanda Askell, Yuntao Bai, Anna Chen, Dawn Drain, Deep Ganguli, Tom Henighan,
  Andy Jones, Nicholas Joseph, Ben Mann, Nova DasSarma, Nelson Elhage, Zac
  Hatfield-Dodds, Danny Hernandez, Jackson Kernion, Kamal Ndousse, Catherine
  Olsson, Dario Amodei, Tom Brown, Jack Clark, Sam McCandlish, Chris Olah, and
  Jared Kaplan. 2021.
\newblock \href {https://doi.org/10.48550/arXiv.2112.00861} {A {General}
  {Language} {Assistant} as a {Laboratory} for {Alignment}}.
\newblock {{arXiv}}:2112.00861 [cs].

\bibitem[{Asprino et~al.(2022)Asprino, Bulla, De~Giorgis, Gangemi, Marinucci,
  and Mongiovi}]{asprinoUncovering2022}
Luigi Asprino, Luana Bulla, Stefano De~Giorgis, Aldo Gangemi, Ludovica
  Marinucci, and Misael Mongiovi. 2022.
\newblock \href {https://doi.org/10.18653/v1/2022.deelio-1.4} {Uncovering
  {Values}: {Detecting} {Latent} {Moral} {Content} from {Natural} {Language}
  with {Explainable} and {Non}-{Trained} {Methods}}.
\newblock In \emph{Proceedings of {Deep} {Learning} {Inside} {Out} ({DeeLIO}
  2022): {The} 3rd {Workshop} on {Knowledge} {Extraction} and {Integration} for
  {Deep} {Learning} {Architectures}}, pages 33--41, Dublin, Ireland and Online.
  Association for Computational Linguistics.

\bibitem[{Bai et~al.(2022{\natexlab{a}})Bai, Jones, Ndousse, Askell, Chen,
  DasSarma, Drain, Fort, Ganguli, Henighan, Joseph, Kadavath, Kernion, Conerly,
  El-Showk, Elhage, Hatfield-Dodds, Hernandez, Hume, Johnston, Kravec, Lovitt,
  Nanda, Olsson, Amodei, Brown, Clark, McCandlish, Olah, Mann, and
  Kaplan}]{baiTraining2022}
Yuntao Bai, Andy Jones, Kamal Ndousse, Amanda Askell, Anna Chen, Nova DasSarma,
  Dawn Drain, Stanislav Fort, Deep Ganguli, Tom Henighan, Nicholas Joseph,
  Saurav Kadavath, Jackson Kernion, Tom Conerly, Sheer El-Showk, Nelson Elhage,
  Zac Hatfield-Dodds, Danny Hernandez, Tristan Hume, Scott Johnston, Shauna
  Kravec, Liane Lovitt, Neel Nanda, Catherine Olsson, Dario Amodei, Tom Brown,
  Jack Clark, Sam McCandlish, Chris Olah, Ben Mann, and Jared Kaplan.
  2022{\natexlab{a}}.
\newblock \href {http://arxiv.org/abs/2204.05862v1} {Training a {Helpful} and
  {Harmless} {Assistant} with {Reinforcement} {Learning} from {Human}
  {Feedback}}.
\newblock {arXiv}: 2204.05862.

\bibitem[{Bai et~al.(2022{\natexlab{b}})Bai, Kadavath, Kundu, Askell, Kernion,
  Jones, Chen, Goldie, Mirhoseini, McKinnon, Chen, Olsson, Olah, Hernandez,
  Drain, Ganguli, Li, Tran-Johnson, Perez, Kerr, Mueller, Ladish, Landau,
  Ndousse, Lukosuite, Lovitt, Sellitto, Elhage, Schiefer, Mercado, DasSarma,
  Lasenby, Larson, Ringer, Johnston, Kravec, Showk, Fort, Lanham,
  Telleen-Lawton, Conerly, Henighan, Hume, Bowman, Hatfield-Dodds, Mann,
  Amodei, Joseph, McCandlish, Brown, and Kaplan}]{baiConstitutional2022}
Yuntao Bai, Saurav Kadavath, Sandipan Kundu, Amanda Askell, Jackson Kernion,
  Andy Jones, Anna Chen, Anna Goldie, Azalia Mirhoseini, Cameron McKinnon,
  Carol Chen, Catherine Olsson, Christopher Olah, Danny Hernandez, Dawn Drain,
  Deep Ganguli, Dustin Li, Eli Tran-Johnson, Ethan Perez, Jamie Kerr, Jared
  Mueller, Jeffrey Ladish, Joshua Landau, Kamal Ndousse, Kamile Lukosuite,
  Liane Lovitt, Michael Sellitto, Nelson Elhage, Nicholas Schiefer, Noemi
  Mercado, Nova DasSarma, Robert Lasenby, Robin Larson, Sam Ringer, Scott
  Johnston, Shauna Kravec, Sheer~El Showk, Stanislav Fort, Tamera Lanham,
  Timothy Telleen-Lawton, Tom Conerly, Tom Henighan, Tristan Hume, Samuel~R.
  Bowman, Zac Hatfield-Dodds, Ben Mann, Dario Amodei, Nicholas Joseph, Sam
  McCandlish, Tom Brown, and Jared Kaplan. 2022{\natexlab{b}}.
\newblock \href {http://arxiv.org/abs/2212.08073v1} {Constitutional {AI}:
  {Harmlessness} from {AI} {Feedback}}.
\newblock {arXiv}: 2212.08073.

\bibitem[{Bakker et~al.(2022)Bakker, Chadwick, Sheahan, Tessler,
  Campbell-Gillingham, Balaguer, McAleese, Glaese, Aslanides, Botvinick, and
  Summerfield}]{bakkerFinetuning2022}
Michiel~A. Bakker, Martin~J. Chadwick, Hannah~R. Sheahan, Michael~Henry
  Tessler, Lucy Campbell-Gillingham, Jan Balaguer, Nat McAleese, Amelia Glaese,
  John Aslanides, Matthew~M. Botvinick, and Christopher Summerfield. 2022.
\newblock \href {http://arxiv.org/abs/2211.15006v1} {Fine-tuning language
  models to find agreement among humans with diverse preferences}.
\newblock {{arXiv}}: 2211.15006v1.

\bibitem[{Bang et~al.(2022)Bang, Yu, Madotto, Lin, Diab, and
  Fung}]{bangEnabling2022}
Yejin Bang, Tiezheng Yu, Andrea Madotto, Zhaojiang Lin, Mona Diab, and Pascale
  Fung. 2022.
\newblock \href {http://arxiv.org/abs/2210.07652v1} {Enabling {Classifiers} to
  {Make} {Judgements} {Explicitly} {Aligned} with {Human} {Values}}.
\newblock {{arXiv}}: 2210.07652v1.

\bibitem[{Bender and Friedman(2018)}]{bender2018data}
Emily~M Bender and Batya Friedman. 2018.
\newblock Data statements for natural language processing: Toward mitigating
  system bias and enabling better science.
\newblock \emph{Transactions of the Association for Computational Linguistics},
  6:587--604.

\bibitem[{Birhane et~al.(2022)Birhane, Isaac, Prabhakaran, Diaz, Elish,
  Gabriel, and Mohamed}]{birhanePower2022}
Abeba Birhane, William Isaac, Vinodkumar Prabhakaran, Mark Diaz,
  Madeleine~Clare Elish, Iason Gabriel, and Shakir Mohamed. 2022.
\newblock \href {https://doi.org/10.1145/3551624.3555290} {Power to the
  {People}? {Opportunities} and {Challenges} for {Participatory} {AI}}.
\newblock In \emph{Equity and {Access} in {Algorithms}, {Mechanisms}, and
  {Optimization}}, {EAAMO} '22, pages 1--8, New York, NY, USA. Association for
  Computing Machinery.

\bibitem[{Bowman et~al.(2022)Bowman, Hyun, Perez, Chen, Pettit, Heiner,
  Lukošiūtė, Askell, Jones, Chen, Goldie, Mirhoseini, McKinnon, Olah,
  Amodei, Amodei, Drain, Li, Tran-Johnson, Kernion, Kerr, Mueller, Ladish,
  Landau, Ndousse, Lovitt, Elhage, Schiefer, Joseph, Mercado, DasSarma, Larson,
  McCandlish, Kundu, Johnston, Kravec, Showk, Fort, Telleen-Lawton, Brown,
  Henighan, Hume, Bai, Hatfield-Dodds, Mann, and Kaplan}]{bowmanMeasuring2022}
Samuel~R. Bowman, Jeeyoon Hyun, Ethan Perez, Edwin Chen, Craig Pettit, Scott
  Heiner, Kamilė Lukošiūtė, Amanda Askell, Andy Jones, Anna Chen, Anna
  Goldie, Azalia Mirhoseini, Cameron McKinnon, Christopher Olah, Daniela
  Amodei, Dario Amodei, Dawn Drain, Dustin Li, Eli Tran-Johnson, Jackson
  Kernion, Jamie Kerr, Jared Mueller, Jeffrey Ladish, Joshua Landau, Kamal
  Ndousse, Liane Lovitt, Nelson Elhage, Nicholas Schiefer, Nicholas Joseph,
  Noemí Mercado, Nova DasSarma, Robin Larson, Sam McCandlish, Sandipan Kundu,
  Scott Johnston, Shauna Kravec, Sheer~El Showk, Stanislav Fort, Timothy
  Telleen-Lawton, Tom Brown, Tom Henighan, Tristan Hume, Yuntao Bai, Zac
  Hatfield-Dodds, Ben Mann, and Jared Kaplan. 2022.
\newblock \href {https://doi.org/10.48550/arXiv.2211.03540} {Measuring
  {Progress} on {Scalable} {Oversight} for {Large} {Language} {Models}}.
\newblock {{arXiv}}:2211.03540 [cs].

\bibitem[{Böhm et~al.(2019)Böhm, Gao, Meyer, Shapira, Dagan, and
  Gurevych}]{bohmBetter2019}
Florian Böhm, Yang Gao, Christian~M. Meyer, Ori Shapira, Ido Dagan, and Iryna
  Gurevych. 2019.
\newblock \href {https://doi.org/10.18653/v1/D19-1307} {Better {Rewards}
  {Yield} {Better} {Summaries}: {Learning} to {Summarise} {Without}
  {References}}.
\newblock In \emph{Proceedings of the 2019 {Conference} on {Empirical}
  {Methods} in {Natural} {Language} {Processing} and the 9th {International}
  {Joint} {Conference} on {Natural} {Language} {Processing}
  ({EMNLP}-{IJCNLP})}, pages 3110--3120, Hong Kong, China. Association for
  Computational Linguistics.

\bibitem[{Campano et~al.(2014)Campano, Durand, and
  Clavel}]{campanoComparative2014}
Sabrina Campano, Jessica Durand, and Chloé Clavel. 2014.
\newblock \href
  {http://www.lrec-conf.org/proceedings/lrec2014/pdf/327\_Paper.pdf}
  {Comparative analysis of verbal alignment in human-human and human-agent
  interactions}.
\newblock In \emph{Proceedings of the {Ninth} {International} {Conference} on
  {Language} {Resources} and {Evaluation} ({LREC}'14)}, pages 4415--4422,
  Reykjavik, Iceland. European Language Resources Association (ELRA).

\bibitem[{Castricato et~al.(2022)Castricato, Havrilla, Matiana, Pieler, Ye,
  Yang, Frazier, and Riedl}]{castricatoRobust2022}
Louis Castricato, Alexander Havrilla, Shahbuland Matiana, Michael Pieler,
  Anbang Ye, Ian Yang, Spencer Frazier, and Mark Riedl. 2022.
\newblock \href {http://arxiv.org/abs/2210.07792v2} {Robust preference learning
  for storytelling via contrastive reinforcement learning}.
\newblock {{arXiv}}: 2210.07792v2 [cs.CL].

\bibitem[{Chang et~al.(2019)Chang, Prabhakaran, and Ordonez}]{chang2019bias}
Kai-Wei Chang, Vinodkumar Prabhakaran, and Vicente Ordonez. 2019.
\newblock Bias and fairness in natural language processing.
\newblock In \emph{Proceedings of the 2019 Conference on Empirical Methods in
  Natural Language Processing and the 9th International Joint Conference on
  Natural Language Processing (EMNLP-IJCNLP): Tutorial Abstracts}.

\bibitem[{Chiesa and Hobbs(2008)}]{chiesa2008making}
Mecca Chiesa and Sandy Hobbs. 2008.
\newblock Making sense of social research: How useful is the hawthorne effect?
\newblock \emph{European Journal of Social Psychology}, 38(1):67--74.

\bibitem[{Christiano et~al.(2017)Christiano, Leike, Brown, Martic, Legg, and
  Amodei}]{christiano2017deep}
Paul~F Christiano, Jan Leike, Tom Brown, Miljan Martic, Shane Legg, and Dario
  Amodei. 2017.
\newblock Deep reinforcement learning from human preferences.
\newblock \emph{Advances in neural information processing systems}, 30.

\bibitem[{Chuang and Schechter(2015)}]{chuang2015stability}
Yating Chuang and Laura Schechter. 2015.
\newblock Stability of experimental and survey measures of risk, time, and
  social preferences: A review and some new results.
\newblock \emph{Journal of development economics}, 117:151--170.

\bibitem[{Davani et~al.(2022)Davani, D{\'\i}az, and
  Prabhakaran}]{davani2022dealing}
Aida~Mostafazadeh Davani, Mark D{\'\i}az, and Vinodkumar Prabhakaran. 2022.
\newblock Dealing with disagreements: Looking beyond the majority vote in
  subjective annotations.
\newblock \emph{Transactions of the Association for Computational Linguistics},
  10:92--110.

\bibitem[{De~Deyne et~al.(2016)De~Deyne, Perfors, and
  Navarro}]{dedeynePredicting2016}
Simon De~Deyne, Amy Perfors, and Daniel~J Navarro. 2016.
\newblock \href {https://aclanthology.org/C16-1175} {Predicting human
  similarity judgments with distributional models: {The} value of word
  associations.}
\newblock In \emph{Proceedings of {COLING} 2016, the 26th {International}
  {Conference} on {Computational} {Linguistics}: {Technical} {Papers}}, pages
  1861--1870, Osaka, Japan. The COLING 2016 Organizing Committee.

\bibitem[{Dell et~al.(2012)Dell, Vaidyanathan, Medhi, Cutrell, and
  Thies}]{dell2012yours}
Nicola Dell, Vidya Vaidyanathan, Indrani Medhi, Edward Cutrell, and William
  Thies. 2012.
\newblock " yours is better!" participant response bias in hci.
\newblock In \emph{Proceedings of the sigchi conference on human factors in
  computing systems}, pages 1321--1330.

\bibitem[{Den~Hengst et~al.(2019)Den~Hengst, Hoogendoorn, Van~Harmelen, and
  Bosman}]{hengstReinforcementLearningPersonalized2019}
Floris Den~Hengst, Mark Hoogendoorn, Frank Van~Harmelen, and Joost Bosman.
  2019.
\newblock Reinforcement learning for personalized dialogue management.
\newblock In \emph{IEEE/WIC/ACM International Conference on Web Intelligence},
  pages 59--67.

\bibitem[{Deng and Poole(2010)}]{deng2010affect}
Liqiong Deng and Marshall~Scott Poole. 2010.
\newblock Affect in web interfaces: A study of the impacts of web page visual
  complexity and order.
\newblock \emph{Mis Quarterly}, pages 711--730.

\bibitem[{Deng et~al.(2022)Deng, Li, Zhang, Ding, and
  Lam}]{dengPersonalized2021}
Yang Deng, Yaliang Li, Wenxuan Zhang, Bolin Ding, and Wai Lam. 2022.
\newblock Toward personalized answer generation in e-commerce via
  multi-perspective preference modeling.
\newblock \emph{ACM Transactions on Information Systems (TOIS)}, 40(4):1--28.

\bibitem[{Derczynski et~al.(2023)Derczynski, Kirk, Balachandran, Kumar,
  Tsvetkov, Leiser, and Mohammad}]{derczynskiAssessing2023}
Leon Derczynski, Hannah~Rose Kirk, Vidhisha Balachandran, Sachin Kumar, Yulia
  Tsvetkov, M.~R. Leiser, and Saif Mohammad. 2023.
\newblock \href {https://doi.org/10.48550/arXiv.2303.18190} {Assessing
  {Language} {Model} {Deployment} with {Risk} {Cards}}.
\newblock {arXiv}:2303.18190 [cs].

\bibitem[{Devlin et~al.(2019)Devlin, Chang, Lee, and
  Toutanova}]{devlinBERT2019}
Jacob Devlin, Ming-Wei Chang, Kenton Lee, and Kristina Toutanova. 2019.
\newblock \href {https://doi.org/10.48550/arXiv.1810.04805} {{BERT}:
  {Pre}-training of {Deep} {Bidirectional} {Transformers} for {Language}
  {Understanding}}.
\newblock {arXiv}:1810.04805 [cs].

\bibitem[{Dhingra et~al.(2017)Dhingra, Li, Li, Gao, Chen, Ahmed, and
  Deng}]{dhingraEndtoEnd2017}
Bhuwan Dhingra, Lihong Li, Xiujun Li, Jianfeng Gao, Yun-Nung Chen, Faisal
  Ahmed, and Li~Deng. 2017.
\newblock \href {https://doi.org/10.18653/v1/P17-1045} {Towards {End}-to-{End}
  {Reinforcement} {Learning} of {Dialogue} {Agents} for {Information}
  {Access}}.
\newblock In \emph{Proceedings of the 55th {Annual} {Meeting} of the
  {Association} for {Computational} {Linguistics} ({Volume} 1: {Long}
  {Papers})}, pages 484--495, Vancouver, Canada. Association for Computational
  Linguistics.

\bibitem[{Dinan et~al.(2019)Dinan, Humeau, Chintagunta, and
  Weston}]{dinanBuild2019}
Emily Dinan, Samuel Humeau, Bharath Chintagunta, and Jason Weston. 2019.
\newblock \href {https://doi.org/10.48550/arXiv.1908.06083} {Build it {Break}
  it {Fix} it for {Dialogue} {Safety}: {Robustness} from {Adversarial} {Human}
  {Attack}}.
\newblock {{arXiv}}:1908.06083 [cs].

\bibitem[{Dong et~al.(2023)Dong, Xiong, Goyal, Zhang, Chow, Pan, Diao, Zhang,
  Shum, and Zhang}]{dongRAFT2023}
Hanze Dong, Wei Xiong, Deepanshu Goyal, Yihan Zhang, Winnie Chow, Rui Pan,
  Shizhe Diao, Jipeng Zhang, Kashun Shum, and Tong Zhang. 2023.
\newblock \href {https://doi.org/10.48550/arXiv.2304.06767} {{{RAFT}}: {{Reward
  rAnked FineTuning}} for {{Generative Foundation Model Alignment}}}.
\newblock {arXiv}: 2304.06767.

\bibitem[{Dubuisson~Duplessis et~al.(2017)Dubuisson~Duplessis, Clavel, and
  Landragin}]{dubuissonduplessisAutomatic2017}
Guillaume Dubuisson~Duplessis, Chloé Clavel, and Frédéric Landragin. 2017.
\newblock \href {https://doi.org/10.18653/v1/W17-5510} {Automatic {Measures} to
  {Characterise} {Verbal} {Alignment} in {Human}-{Agent} {Interaction}}.
\newblock In \emph{Proceedings of the 18th {Annual} {SIGdial} {Meeting} on
  {Discourse} and {Dialogue}}, pages 71--81, Saarbrücken, Germany. Association
  for Computational Linguistics.

\bibitem[{Ficler and Goldberg(2017)}]{ficlerControlling2017}
Jessica Ficler and Yoav Goldberg. 2017.
\newblock \href {https://doi.org/10.18653/v1/W17-4912} {Controlling
  {Linguistic} {Style} {Aspects} in {Neural} {Language} {Generation}}.
\newblock In \emph{Proceedings of the {Workshop} on {Stylistic} {Variation}},
  pages 94--104, Copenhagen, Denmark. Association for Computational
  Linguistics.

\bibitem[{Fischer(2017)}]{fischerPersonalityValuesCulture2017}
Ronald Fischer. 2017.
\newblock Personality, {{Values}}, {{Culture}}.
\newblock In \emph{Personality, {{Values}}, {{Culture}}: {{An Evolutionary
  Approach}}}, Culture and {{Psychology}}, pages i--ii. {Cambridge University
  Press}, {Cambridge}.

\bibitem[{Forbes et~al.(2020)Forbes, Hwang, Shwartz, Sap, and
  Choi}]{forbesSocial2020}
Maxwell Forbes, Jena~D. Hwang, Vered Shwartz, Maarten Sap, and Yejin Choi.
  2020.
\newblock \href {https://doi.org/10.18653/v1/2020.emnlp-main.48} {Social
  {Chemistry} 101: {Learning} to {Reason} about {Social} and {Moral} {Norms}}.
\newblock In \emph{Proceedings of the 2020 {Conference} on {Empirical}
  {Methods} in {Natural} {Language} {Processing} ({EMNLP})}, pages 653--670,
  Online. Association for Computational Linguistics.

\bibitem[{Friedman et~al.(1994)Friedman, Herskovitz, and
  Pollack}]{friedman1994biasing}
Hershey~H Friedman, Paul~J Herskovitz, and Simcha Pollack. 1994.
\newblock The biasing effects of scale-checking styles on response to a likert
  scale.
\newblock In \emph{Proceedings of the American statistical association annual
  conference: survey research methods}, volume 792, pages 792--795.

\bibitem[{Futrell and Levy(2019)}]{futrellRNNs2019}
Richard Futrell and Roger~P. Levy. 2019.
\newblock \href {https://doi.org/10.7275/jb34-9986} {Do {RNNs} learn human-like
  abstract word order preferences?}
\newblock In \emph{Proceedings of the {Society} for {Computation} in
  {Linguistics} ({SCiL}) 2019}, pages 50--59.

\bibitem[{Ganguli et~al.(2022)Ganguli, Lovitt, Kernion, Askell, Bai, Kadavath,
  Mann, Perez, Schiefer, Ndousse, Jones, Bowman, Chen, Conerly, DasSarma,
  Drain, Elhage, El-Showk, Fort, Hatfield-Dodds, Henighan, Hernandez, Hume,
  Jacobson, Johnston, Kravec, Olsson, Ringer, Tran-Johnson, Amodei, Brown,
  Joseph, McCandlish, Olah, Kaplan, and Clark}]{ganguliRed2022}
Deep Ganguli, Liane Lovitt, Jackson Kernion, Amanda Askell, Yuntao Bai, Saurav
  Kadavath, Ben Mann, Ethan Perez, Nicholas Schiefer, Kamal Ndousse, Andy
  Jones, Sam Bowman, Anna Chen, Tom Conerly, Nova DasSarma, Dawn Drain, Nelson
  Elhage, Sheer El-Showk, Stanislav Fort, Zac Hatfield-Dodds, Tom Henighan,
  Danny Hernandez, Tristan Hume, Josh Jacobson, Scott Johnston, Shauna Kravec,
  Catherine Olsson, Sam Ringer, Eli Tran-Johnson, Dario Amodei, Tom Brown,
  Nicholas Joseph, Sam McCandlish, Chris Olah, Jared Kaplan, and Jack Clark.
  2022.
\newblock \href {http://arxiv.org/abs/2209.07858v2} {Red {Teaming} {Language}
  {Models} to {Reduce} {Harms}: {Methods}, {Scaling} {Behaviors}, and {Lessons}
  {Learned}}.
\newblock {{arXiv}}: 2209.07858v2.

\bibitem[{Gao et~al.(2020)Gao, Zhang, Galley, Brockett, and
  Dolan}]{gaoDialogue2020}
Xiang Gao, Yizhe Zhang, Michel Galley, Chris Brockett, and Bill Dolan. 2020.
\newblock \href {https://doi.org/10.18653/v1/2020.emnlp-main.28} {Dialogue
  {Response} {Ranking} {Training} with {Large}-{Scale} {Human} {Feedback}
  {Data}}.
\newblock In \emph{Proceedings of the 2020 {Conference} on {Empirical}
  {Methods} in {Natural} {Language} {Processing} ({EMNLP})}, pages 386--395,
  Online. Association for Computational Linguistics.

\bibitem[{Gao et~al.(2018)Gao, Meyer, and Gurevych}]{gaoAPRIL2018}
Yang Gao, Christian~M. Meyer, and Iryna Gurevych. 2018.
\newblock \href {https://doi.org/10.18653/v1/D18-1445} {{APRIL}:
  {Interactively} learning to summarise by combining active preference learning
  and reinforcement learning}.
\newblock In \emph{Proceedings of the 2018 conference on empirical methods in
  natural language processing}, pages 4120--4130, Brussels, Belgium.
  Association for Computational Linguistics.

\bibitem[{Gehman et~al.(2020)Gehman, Gururangan, Sap, Choi, and
  Smith}]{gehmanRealToxicityPromptsEvaluatingNeural2020a}
Samuel Gehman, Suchin Gururangan, Maarten Sap, Yejin Choi, and Noah~A. Smith.
  2020.
\newblock \href {https://doi.org/10.18653/v1/2020.findings-emnlp.301}
  {{{RealToxicityPrompts}}: {{Evaluating Neural Toxic Degeneration}} in
  {{Language Models}}}.
\newblock In \emph{Findings of the {{Association}} for {{Computational
  Linguistics}}: {{EMNLP}} 2020}, pages 3356--3369, {Online}. {Association for
  Computational Linguistics}.

\bibitem[{Geva et~al.(2019)Geva, Goldberg, and Berant}]{geva2019we}
Mor Geva, Yoav Goldberg, and Jonathan Berant. 2019.
\newblock Are we modeling the task or the annotator? an investigation of
  annotator bias in natural language understanding datasets.
\newblock In \emph{Proceedings of the 2019 Conference on Empirical Methods in
  Natural Language Processing and the 9th International Joint Conference on
  Natural Language Processing (EMNLP-IJCNLP)}, pages 1161--1166.

\bibitem[{Glaese et~al.(2022)Glaese, McAleese, Trębacz, Aslanides, Firoiu,
  Ewalds, Rauh, Weidinger, Chadwick, Thacker, Campbell-Gillingham, Uesato,
  Huang, Comanescu, Yang, See, Dathathri, Greig, Chen, Fritz, Elias, Green,
  Mokrá, Fernando, Wu, Foley, Young, Gabriel, Isaac, Mellor, Hassabis,
  Kavukcuoglu, Hendricks, and Irving}]{glaeseImproving2022}
Amelia Glaese, Nat McAleese, Maja Trębacz, John Aslanides, Vlad Firoiu, Timo
  Ewalds, Maribeth Rauh, Laura Weidinger, Martin Chadwick, Phoebe Thacker, Lucy
  Campbell-Gillingham, Jonathan Uesato, Po-Sen Huang, Ramona Comanescu, Fan
  Yang, Abigail See, Sumanth Dathathri, Rory Greig, Charlie Chen, Doug Fritz,
  Jaume~Sanchez Elias, Richard Green, Soňa Mokrá, Nicholas Fernando, Boxi Wu,
  Rachel Foley, Susannah Young, Iason Gabriel, William Isaac, John Mellor,
  Demis Hassabis, Koray Kavukcuoglu, Lisa~Anne Hendricks, and Geoffrey Irving.
  2022.
\newblock \href {http://arxiv.org/abs/2209.14375v1} {Improving alignment of
  dialogue agents via targeted human judgements}.
\newblock {{arXiv}}: 2209.14375v1.

\bibitem[{Gordon et~al.(2022)Gordon, Lam, Park, Patel, Hancock, Hashimoto, and
  Bernstein}]{gordon2022jury}
Mitchell~L Gordon, Michelle~S Lam, Joon~Sung Park, Kayur Patel, Jeff Hancock,
  Tatsunori Hashimoto, and Michael~S Bernstein. 2022.
\newblock Jury learning: Integrating dissenting voices into machine learning
  models.
\newblock In \emph{Proceedings of the 2022 CHI Conference on Human Factors in
  Computing Systems}, pages 1--19.

\bibitem[{Gros et~al.(2021)Gros, Li, and Yu}]{grosRUARobot2021}
David Gros, Yu~Li, and Zhou Yu. 2021.
\newblock \href {https://doi.org/10.18653/v1/2021.acl-long.544} {The
  {R}-{U}-{A}-{Robot} {Dataset}: {Helping} {Avoid} {Chatbot} {Deception} by
  {Detecting} {User} {Questions} {About} {Human} or {Non}-{Human} {Identity}}.
\newblock In \emph{Proceedings of the 59th {Annual} {Meeting} of the
  {Association} for {Computational} {Linguistics} and the 11th {International}
  {Joint} {Conference} on {Natural} {Language} {Processing} ({Volume} 1: {Long}
  {Papers})}, pages 6999--7013, Online. Association for Computational
  Linguistics.

\bibitem[{Hancock et~al.(2019)Hancock, Bordes, Mazare, and
  Weston}]{hancockLearning2019}
Braden Hancock, Antoine Bordes, Pierre-Emmanuel Mazare, and Jason Weston. 2019.
\newblock \href {https://doi.org/10.18653/v1/P19-1358} {Learning from
  {Dialogue} after {Deployment}: {Feed} {Yourself}, {Chatbot}!}
\newblock In \emph{Proceedings of the 57th {Annual} {Meeting} of the
  {Association} for {Computational} {Linguistics}}, pages 3667--3684, Florence,
  Italy. Association for Computational Linguistics.

\bibitem[{Hendrycks et~al.(2020)Hendrycks, Burns, Basart, Critch, Li, Song, and
  Steinhardt}]{hendrycksAligning2020}
Dan Hendrycks, Collin Burns, Steven Basart, Andrew Critch, Jerry Li, Dawn Song,
  and Jacob Steinhardt. 2020.
\newblock \href {http://arxiv.org/abs/2008.02275v5} {Aligning {AI} {With}
  {Shared} {Human} {Values}}.
\newblock {{arXiv}}: 2008.02275v5.

\bibitem[{Hochreiter and Schmidhuber(1997)}]{hochreiterLongShorttermMemory1997}
Sepp Hochreiter and J{\"u}rgen Schmidhuber. 1997.
\newblock \href {https://doi.org/10.1162/neco.1997.9.8.1735} {Long {{Short-term
  Memory}}}.
\newblock \emph{Neural computation}, 9:1735--80.

\bibitem[{Honovich et~al.(2022)Honovich, Scialom, Levy, and
  Schick}]{honovichUnnatural2022}
Or~Honovich, Thomas Scialom, Omer Levy, and Timo Schick. 2022.
\newblock \href {http://arxiv.org/abs/2212.09689v1} {Unnatural {Instructions}:
  {Tuning} {Language} {Models} with ({Almost}) {No} {Human} {Labor}}.
\newblock {{arXiv}}: 2212.09689v1.

\bibitem[{Hoover et~al.(2020)Hoover, Portillo-Wightman, Yeh, Havaldar, Davani,
  Lin, Kennedy, Atari, Kamel, Mendlen, Moreno, Park, Chang, Chin, Leong, Leung,
  Mirinjian, and Dehghani}]{hooverMoral2020}
Joe Hoover, Gwenyth Portillo-Wightman, Leigh Yeh, Shreya Havaldar,
  Aida~Mostafazadeh Davani, Ying Lin, Brendan Kennedy, Mohammad Atari, Zahra
  Kamel, Madelyn Mendlen, Gabriela Moreno, Christina Park, Tingyee~E. Chang,
  Jenna Chin, Christian Leong, Jun~Yen Leung, Arineh Mirinjian, and Morteza
  Dehghani. 2020.
\newblock \href {https://doi.org/10.1177/1948550619876629} {Moral {Foundations}
  {Twitter} {Corpus}: {A} {Collection} of 35k {Tweets} {Annotated} for {Moral}
  {Sentiment}}.
\newblock \emph{Social Psychological and Personality Science},
  11(8):1057--1071.
\newblock Publisher: SAGE Publications Inc.

\bibitem[{Hosking et~al.(2023)Hosking, Blunsom, and Bartolo}]{hoskingHuman2023}
Tom Hosking, Phil Blunsom, and Max Bartolo. 2023.
\newblock \href {https://doi.org/10.48550/arXiv.2309.16349} {Human {{Feedback}}
  is not {{Gold Standard}}}.
\newblock {arXiv}: 2309.16349.

\bibitem[{Hsieh and Kocielnik(2016)}]{hsieh2016you}
Gary Hsieh and Rafa{\l} Kocielnik. 2016.
\newblock You get who you pay for: The impact of incentives on participation
  bias.
\newblock In \emph{Proceedings of the 19th ACM conference on computer-supported
  cooperative work \& social computing}, pages 823--835.

\bibitem[{Huang et~al.(2022)Huang, Wormley, and Cohen}]{huangLearning2022}
Xiaolei Huang, Alexandra Wormley, and Adam Cohen. 2022.
\newblock \href {https://doi.org/10.1145/3511095.3531269} {Learning to {Adapt}
  {Domain} {Shifts} of {Moral} {Values} via {Instance} {Weighting}}.
\newblock {{arXiv}}: 2204.07603v2.

\bibitem[{Jaques et~al.(2019)Jaques, Ghandeharioun, Shen, Ferguson, Lapedriza,
  Jones, Gu, and Picard}]{jaquesWay2019}
Natasha Jaques, Asma Ghandeharioun, Judy~Hanwen Shen, Craig Ferguson, Agata
  Lapedriza, Noah Jones, Shixiang Gu, and Rosalind Picard. 2019.
\newblock \href {https://doi.org/10.48550/arXiv.1907.00456} {Way {Off}-{Policy}
  {Batch} {Deep} {Reinforcement} {Learning} of {Implicit} {Human} {Preferences}
  in {Dialog}}.
\newblock {{arXiv}}:1907.00456 [cs, stat].

\bibitem[{Jaques et~al.(2020)Jaques, Shen, Ghandeharioun, Ferguson, Lapedriza,
  Jones, Gu, and Picard}]{jaquesHumancentric2020}
Natasha Jaques, Judy~Hanwen Shen, Asma Ghandeharioun, Craig Ferguson, Agata
  Lapedriza, Noah Jones, Shixiang Gu, and Rosalind Picard. 2020.
\newblock \href {https://doi.org/10.18653/v1/2020.emnlp-main.327}
  {Human-centric dialog training via offline reinforcement learning}.
\newblock In \emph{Proceedings of the 2020 conference on empirical methods in
  natural language processing ({EMNLP})}, pages 3985--4003, Online. Association
  for Computational Linguistics.

\bibitem[{Jentzsch et~al.(2019)Jentzsch, Schramowski, Rothkopf, and
  Kersting}]{jentzschSemantics2019}
Sophie Jentzsch, Patrick Schramowski, Constantin Rothkopf, and Kristian
  Kersting. 2019.
\newblock \href {https://doi.org/10.1145/3306618.3314267} {Semantics {Derived}
  {Automatically} from {Language} {Corpora} {Contain} {Human}-like {Moral}
  {Choices}}.
\newblock In \emph{Proceedings of the 2019 {AAAI}/{ACM} {Conference} on {AI},
  {Ethics}, and {Society}}, {AIES} '19, pages 37--44, New York, NY, USA.
  Association for Computing Machinery.

\bibitem[{Ji et~al.(2022)Ji, Lee, Frieske, Yu, Su, Xu, Ishii, Bang, Madotto,
  and Fung}]{jiSurveyHallucinationNatural2022}
Ziwei Ji, Nayeon Lee, Rita Frieske, Tiezheng Yu, Dan Su, Yan Xu, Etsuko Ishii,
  Yejin Bang, Andrea Madotto, and Pascale Fung. 2022.
\newblock \href {https://doi.org/10.48550/arXiv.2202.03629} {Survey of
  {{Hallucination}} in {{Natural Language Generation}}}.

\bibitem[{Jiang et~al.(2022)Jiang, Hwang, Bhagavatula, Bras, Liang, Dodge,
  Sakaguchi, Forbes, Borchardt, Gabriel, Tsvetkov, Etzioni, Sap, Rini, and
  Choi}]{jiangCan2022}
Liwei Jiang, Jena~D. Hwang, Chandra Bhagavatula, Ronan~Le Bras, Jenny Liang,
  Jesse Dodge, Keisuke Sakaguchi, Maxwell Forbes, Jon Borchardt, Saadia
  Gabriel, Yulia Tsvetkov, Oren Etzioni, Maarten Sap, Regina Rini, and Yejin
  Choi. 2022.
\newblock \href {https://doi.org/10.48550/arXiv.2110.07574} {Can {Machines}
  {Learn} {Morality}? {The} {Delphi} {Experiment}}.
\newblock {{arXiv}}:2110.07574 [cs].

\bibitem[{Jin et~al.(2022)Jin, Levine, Gonzalez, Kamal, Sap, Sachan, Mihalcea,
  Tenenbaum, and Schölkopf}]{jinWhen2022}
Zhijing Jin, Sydney Levine, Fernando Gonzalez, Ojasv Kamal, Maarten Sap,
  Mrinmaya Sachan, Rada Mihalcea, Josh Tenenbaum, and Bernhard Schölkopf.
  2022.
\newblock \href {http://arxiv.org/abs/2210.01478v3} {When to {Make}
  {Exceptions}: {Exploring} {Language} {Models} as {Accounts} of {Human}
  {Moral} {Judgment}}.
\newblock {{arXiv}}: 2210.01478v3.

\bibitem[{Jobin et~al.(2019)Jobin, Ienca, and Vayena}]{jobin2019global}
Anna Jobin, Marcello Ienca, and Effy Vayena. 2019.
\newblock The global landscape of ai ethics guidelines.
\newblock \emph{Nature Machine Intelligence}, 1(9):389--399.

\bibitem[{Ju et~al.(2022)Ju, Xu, Boureau, and Weston}]{juLearning2022}
Da~Ju, Jing Xu, Y.-Lan Boureau, and Jason Weston. 2022.
\newblock \href {https://doi.org/10.48550/arXiv.2208.03295} {Learning from data
  in the mixed adversarial non-adversarial case: {Finding} the helpers and
  ignoring the trolls}.
\newblock {{arXiv}}:2208.03295 [cs].

\bibitem[{Kiesel et~al.(2022)Kiesel, Alshomary, Handke, Cai, Wachsmuth, and
  Stein}]{kieselIdentifying2022}
Johannes Kiesel, Milad Alshomary, Nicolas Handke, Xiaoni Cai, Henning
  Wachsmuth, and Benno Stein. 2022.
\newblock \href {https://doi.org/10.18653/v1/2022.acl-long.306} {Identifying
  the {Human} {Values} behind {Arguments}}.
\newblock In \emph{Proceedings of the 60th {Annual} {Meeting} of the
  {Association} for {Computational} {Linguistics} ({Volume} 1: {Long}
  {Papers})}, pages 4459--4471, Dublin, Ireland. Association for Computational
  Linguistics.

\bibitem[{Kirk et~al.(2021)Kirk, Jun, Volpin, Iqbal, Benussi, Dreyer,
  Shtedritski, and Asano}]{kirkBiasOutoftheBoxEmpirical2021}
Hannah~Rose Kirk, Yennie Jun, Filippo Volpin, Haider Iqbal, Elias Benussi,
  Frederic Dreyer, Aleksandar Shtedritski, and Yuki Asano. 2021.
\newblock Bias {{Out-of-the-Box}}: {{An Empirical Analysis}} of
  {{Intersectional Occupational Biases}} in {{Popular Generative Language
  Models}}.
\newblock In \emph{Advances in {{Neural Information Processing Systems}}},
  volume~34, pages 2611--2624.

\bibitem[{Kirk et~al.(2023{\natexlab{a}})Kirk, Vidgen, R{\"o}ttger, and
  Hale}]{kirkEmpty2023b}
Hannah~Rose Kirk, Bertie Vidgen, Paul R{\"o}ttger, and Scott~A. Hale.
  2023{\natexlab{a}}.
\newblock \href {http://arxiv.org/abs/2310.02457} {The {{Empty Signifier
  Problem}}: {{Towards Clearer Paradigms}} for {{Operationalising}}
  "{{Alignment}}" in {{Large Language Models}}}.
\newblock {arXiv}: 2310.02457.

\bibitem[{Kirk et~al.(2023{\natexlab{b}})Kirk, Vidgen, R{\"o}ttger, and
  Hale}]{kirk2023personalisation}
Hannah~Rose Kirk, Bertie Vidgen, Paul R{\"o}ttger, and Scott~A. Hale.
  2023{\natexlab{b}}.
\newblock \href {https://doi.org/10.48550/arXiv.2303.05453} {Personalisation
  within bounds: {{A}} risk taxonomy and policy framework for the alignment of
  large language models with personalised feedback}.
\newblock {arXiv}:2303.05453.

\bibitem[{Korbak et~al.(2023)Korbak, Shi, Chen, Bhalerao, Buckley, Phang,
  Bowman, and Perez}]{korbakPretraining2023}
Tomasz Korbak, Kejian Shi, Angelica Chen, Rasika Bhalerao, Christopher~L.
  Buckley, Jason Phang, Samuel~R. Bowman, and Ethan Perez. 2023.
\newblock \href {https://doi.org/10.48550/arXiv.2302.08582} {Pretraining
  {Language} {Models} with {Human} {Preferences}}.
\newblock {{arXiv}}:2302.08582 [cs].

\bibitem[{Kreutzer et~al.(2017)Kreutzer, Sokolov, and
  Riezler}]{kreutzerBandit2017}
Julia Kreutzer, Artem Sokolov, and Stefan Riezler. 2017.
\newblock \href {https://doi.org/10.18653/v1/P17-1138} {Bandit {Structured}
  {Prediction} for {Neural} {Sequence}-to-{Sequence} {Learning}}.
\newblock In \emph{Proceedings of the 55th {Annual} {Meeting} of the
  {Association} for {Computational} {Linguistics} ({Volume} 1: {Long}
  {Papers})}, pages 1503--1513, Vancouver, Canada. Association for
  Computational Linguistics.

\bibitem[{Kreutzer et~al.(2018)Kreutzer, Uyheng, and
  Riezler}]{kreutzerReliability2018}
Julia Kreutzer, Joshua Uyheng, and Stefan Riezler. 2018.
\newblock \href {https://doi.org/10.18653/v1/P18-1165} {Reliability and
  {Learnability} of {Human} {Bandit} {Feedback} for {Sequence}-to-{Sequence}
  {Reinforcement} {Learning}}.
\newblock In \emph{Proceedings of the 56th {Annual} {Meeting} of the
  {Association} for {Computational} {Linguistics} ({Volume} 1: {Long}
  {Papers})}, pages 1777--1788, Melbourne, Australia. Association for
  Computational Linguistics.

\bibitem[{Laclau(2005)}]{laclauPopulist2005}
Ernesto Laclau. 2005.
\newblock \emph{On populist reason}.
\newblock Verso, London New York (N.Y.).

\bibitem[{Lacy(2001)}]{lacy2001theory}
Dean Lacy. 2001.
\newblock A theory of nonseparable preferences in survey responses.
\newblock \emph{American Journal of Political Science}, pages 239--258.

\bibitem[{Lawrence and Riezler(2018)}]{lawrenceImproving2018}
Carolin Lawrence and Stefan Riezler. 2018.
\newblock \href {https://doi.org/10.18653/v1/P18-1169} {Improving a {Neural}
  {Semantic} {Parser} by {Counterfactual} {Learning} from {Human} {Bandit}
  {Feedback}}.
\newblock In \emph{Proceedings of the 56th {Annual} {Meeting} of the
  {Association} for {Computational} {Linguistics} ({Volume} 1: {Long}
  {Papers})}, pages 1820--1830, Melbourne, Australia. Association for
  Computational Linguistics.

\bibitem[{Lawrence et~al.(2017)Lawrence, Sokolov, and
  Riezler}]{lawrenceCounterfactual2017}
Carolin Lawrence, Artem Sokolov, and Stefan Riezler. 2017.
\newblock \href {https://doi.org/10.18653/v1/D17-1272} {Counterfactual
  {Learning} from {Bandit} {Feedback} under {Deterministic} {Logging} : {A}
  {Case} {Study} in {Statistical} {Machine} {Translation}}.
\newblock In \emph{Proceedings of the 2017 {Conference} on {Empirical}
  {Methods} in {Natural} {Language} {Processing}}, pages 2566--2576,
  Copenhagen, Denmark. Association for Computational Linguistics.

\bibitem[{Lee et~al.(2009)Lee, Amir, and Ariely}]{lee2009search}
Leonard Lee, On~Amir, and Dan Ariely. 2009.
\newblock In search of homo economicus: Cognitive noise and the role of emotion
  in preference consistency.
\newblock \emph{Journal of consumer research}, 36(2):173--187.

\bibitem[{Li et~al.(2017{\natexlab{a}})Li, Miller, Chopra, Ranzato, and
  Weston}]{liDialogue2017}
Jiwei Li, Alexander~H. Miller, Sumit Chopra, Marc'Aurelio Ranzato, and Jason
  Weston. 2017{\natexlab{a}}.
\newblock \href {https://doi.org/10.48550/arXiv.1611.09823} {Dialogue
  {Learning} {With} {Human}-{In}-{The}-{Loop}}.
\newblock {{arXiv}}:1611.09823 [cs].

\bibitem[{Li et~al.(2017{\natexlab{b}})Li, Miller, Chopra, Ranzato, and
  Weston}]{liLearning2017}
Jiwei Li, Alexander~H. Miller, Sumit Chopra, Marc'Aurelio Ranzato, and Jason
  Weston. 2017{\natexlab{b}}.
\newblock \href {https://doi.org/10.48550/arXiv.1612.04936} {Learning through
  {Dialogue} {Interactions} by {Asking} {Questions}}.
\newblock {{arXiv}}:1612.04936 [cs].

\bibitem[{Li et~al.(2016)Li, Monroe, Ritter, Jurafsky, Galley, and
  Gao}]{liDeep2016}
Jiwei Li, Will Monroe, Alan Ritter, Dan Jurafsky, Michel Galley, and Jianfeng
  Gao. 2016.
\newblock \href {https://doi.org/10.18653/v1/D16-1127} {Deep {Reinforcement}
  {Learning} for {Dialogue} {Generation}}.
\newblock In \emph{Proceedings of the 2016 {Conference} on {Empirical}
  {Methods} in {Natural} {Language} {Processing}}, pages 1192--1202, Austin,
  Texas. Association for Computational Linguistics.

\bibitem[{Li et~al.(2019)Li, Lin, Hoover, Whitehead, Voss, Dehghani, and
  Ji}]{liMultilingual2019}
Manling Li, Ying Lin, Joseph Hoover, Spencer Whitehead, Clare Voss, Morteza
  Dehghani, and Heng Ji. 2019.
\newblock \href {https://doi.org/10.18653/v1/N19-4019} {Multilingual {Entity},
  {Relation}, {Event} and {Human} {Value} {Extraction}}.
\newblock In \emph{Proceedings of the 2019 {Conference} of the {North}
  {American} {Chapter} of the {Association} for {Computational} {Linguistics}
  ({Demonstrations})}, pages 110--115, Minneapolis, Minnesota. Association for
  Computational Linguistics.

\bibitem[{Lin et~al.(2022)Lin, Hilton, and
  Evans}]{linTruthfulQAMeasuringHow2022}
Stephanie Lin, Jacob Hilton, and Owain Evans. 2022.
\newblock \href {https://doi.org/10.18653/v1/2022.acl-long.229}
  {{{TruthfulQA}}: {{Measuring How Models Mimic Human Falsehoods}}}.
\newblock In \emph{Proceedings of the 60th {{Annual Meeting}} of the
  {{Association}} for {{Computational Linguistics}} ({{Volume}} 1: {{Long
  Papers}})}, pages 3214--3252, {Dublin, Ireland}. {Association for
  Computational Linguistics}.

\bibitem[{Lin et~al.(2017)Lin, Hoover, Dehghani, Mooijman, and
  Ji}]{linAcquiring2017}
Ying Lin, Joe Hoover, Morteza Dehghani, Marlon Mooijman, and Heng Ji. 2017.
\newblock \href {http://arxiv.org/abs/1709.05467v1} {Acquiring {Background}
  {Knowledge} to {Improve} {Moral} {Value} {Prediction}}.
\newblock {{arXiv}}: 1709.05467v1.

\bibitem[{Liu et~al.(2021{\natexlab{a}})Liu, Sap, Lu, Swayamdipta, Bhagavatula,
  Smith, and Choi}]{liuDExperts2021}
Alisa Liu, Maarten Sap, Ximing Lu, Swabha Swayamdipta, Chandra Bhagavatula,
  Noah~A. Smith, and Yejin Choi. 2021{\natexlab{a}}.
\newblock \href {https://doi.org/10.18653/v1/2021.acl-long.522} {{DExperts}:
  {Decoding}-{Time} {Controlled} {Text} {Generation} with {Experts} and
  {Anti}-{Experts}}.
\newblock In \emph{Proceedings of the 59th {Annual} {Meeting} of the
  {Association} for {Computational} {Linguistics} and the 11th {International}
  {Joint} {Conference} on {Natural} {Language} {Processing} ({Volume} 1: {Long}
  {Papers})}, pages 6691--6706, Online. Association for Computational
  Linguistics.

\bibitem[{Liu et~al.(2018)Liu, Tür, Hakkani-Tür, Shah, and
  Heck}]{liuDialogue2018}
Bing Liu, Gokhan Tür, Dilek Hakkani-Tür, Pararth Shah, and Larry Heck. 2018.
\newblock \href {https://doi.org/10.18653/v1/N18-1187} {Dialogue {Learning}
  with {Human} {Teaching} and {Feedback} in {End}-to-{End} {Trainable}
  {Task}-{Oriented} {Dialogue} {Systems}}.
\newblock In \emph{Proceedings of the 2018 {Conference} of the {North}
  {American} {Chapter} of the {Association} for {Computational} {Linguistics}:
  {Human} {Language} {Technologies}, {Volume} 1 ({Long} {Papers})}, pages
  2060--2069, New Orleans, Louisiana. Association for Computational
  Linguistics.

\bibitem[{Liu et~al.(2023{\natexlab{a}})Liu, Sferrazza, and
  Abbeel}]{liuChain2023}
Hao Liu, Carmelo Sferrazza, and Pieter Abbeel. 2023{\natexlab{a}}.
\newblock \href {https://doi.org/10.48550/arXiv.2302.02676} {Chain of
  {{Hindsight Aligns Language Models}} with {{Feedback}}}.
\newblock {arXiv}: 2302.02676.

\bibitem[{Liu et~al.(2023{\natexlab{b}})Liu, Sferrazza, and
  Abbeel}]{liuLanguages2023}
Hao Liu, Carmelo Sferrazza, and Pieter Abbeel. 2023{\natexlab{b}}.
\newblock \href {http://arxiv.org/abs/2302.02676v2} {Languages are rewards:
  {Chain} of hindsight finetuning using human feedback}.
\newblock {{arXiv}}: 2302.02676v2 [cs.LG].

\bibitem[{Liu et~al.(2021{\natexlab{b}})Liu, Jia, Wei, Xu, Wang, and
  Vosoughi}]{liuMitigating2021}
Ruibo Liu, Chenyan Jia, Jason Wei, Guangxuan Xu, Lili Wang, and Soroush
  Vosoughi. 2021{\natexlab{b}}.
\newblock \href {https://doi.org/10.1609/aaai.v35i17.17744} {Mitigating
  {Political} {Bias} in {Language} {Models} through {Reinforced}
  {Calibration}}.
\newblock \emph{Proceedings of the AAAI Conference on Artificial Intelligence},
  35(17):14857--14866.
\newblock Number: 17.

\bibitem[{Liu et~al.(2023{\natexlab{c}})Liu, Jia, Zhang, Zhuang, Liu, and
  Vosoughi}]{liuSecond2023}
Ruibo Liu, Chenyan Jia, Ge~Zhang, Ziyu Zhuang, Tony~X. Liu, and Soroush
  Vosoughi. 2023{\natexlab{c}}.
\newblock \href {http://arxiv.org/abs/2301.00355v2} {Second {Thoughts} are
  {Best}: {Learning} to {Re}-{Align} {With} {Human} {Values} from {Text}
  {Edits}}.
\newblock {{arXiv}}: 2301.00355v2.

\bibitem[{Liu et~al.(2023{\natexlab{d}})Liu, Yang, Jia, Zhang, Zhou, Dai, Yang,
  and Vosoughi}]{liuTraining2023}
Ruibo Liu, Ruixin Yang, Chenyan Jia, Ge~Zhang, Denny Zhou, Andrew~M. Dai, Diyi
  Yang, and Soroush Vosoughi. 2023{\natexlab{d}}.
\newblock \href {https://doi.org/10.48550/arXiv.2305.16960} {Training
  {{Socially Aligned Language Models}} in {{Simulated Human Society}}}.
\newblock {arXiv}: 2305.16960.

\bibitem[{Liu et~al.(2022)Liu, Zhang, Feng, and Vosoughi}]{liuAligning2022}
Ruibo Liu, Ge~Zhang, Xinyu Feng, and Soroush Vosoughi. 2022.
\newblock \href {https://doi.org/10.18653/v1/2022.findings-naacl.18} {Aligning
  {Generative} {Language} {Models} with {Human} {Values}}.
\newblock In \emph{Findings of the {Association} for {Computational}
  {Linguistics}: {NAACL} 2022}, pages 241--252, Seattle, United States.
  Association for Computational Linguistics.

\bibitem[{Lourie et~al.(2021)Lourie, Bras, and Choi}]{lourieSCRUPLES2021}
Nicholas Lourie, Ronan~Le Bras, and Yejin Choi. 2021.
\newblock \href {https://doi.org/10.1609/aaai.v35i15.17589} {{SCRUPLES}: {A}
  {Corpus} of {Community} {Ethical} {Judgments} on 32,000 {Real}-{Life}
  {Anecdotes}}.
\newblock \emph{Proceedings of the AAAI Conference on Artificial Intelligence},
  35(15):13470--13479.
\newblock Number: 15.

\bibitem[{Lu et~al.(2022)Lu, Bao, He, Wang, Wu, and Wang}]{luBoosting2022}
Hua Lu, Siqi Bao, Huang He, Fan Wang, Hua Wu, and Haifeng Wang. 2022.
\newblock \href {http://arxiv.org/abs/2208.14165v1} {Towards {Boosting} the
  {Open}-{Domain} {Chatbot} with {Human} {Feedback}}.
\newblock {{arXiv}}: 2208.14165v1.

\bibitem[{Lucy and Bamman(2021)}]{lucyGenderRepresentationBias2021}
Li~Lucy and David Bamman. 2021.
\newblock \href {https://doi.org/10.18653/v1/2021.nuse-1.5} {Gender and
  {{Representation Bias}} in {{GPT-3 Generated Stories}}}.
\newblock In \emph{Proceedings of the {{Third Workshop}} on {{Narrative
  Understanding}}}, pages 48--55, {Virtual}. {Association for Computational
  Linguistics}.

\bibitem[{Lévi-Strauss(1987)}]{levi-straussIntroduction1987}
Claude Lévi-Strauss. 1987.
\newblock \emph{Introduction to the work of {Marcel} {Mauss}}.
\newblock Routledge \& Kegan Paul, London.

\bibitem[{Maeda(2015)}]{maeda2015response}
Hotaka Maeda. 2015.
\newblock Response option configuration of online administered likert scales.
\newblock \emph{International Journal of Social Research Methodology},
  18(1):15--26.

\bibitem[{Maheshwari et~al.(2017)Maheshwari, Reganti, Gupta, Jamatia, Kumar,
  Gambäck, and Das}]{maheshwariSocietal2017}
Tushar Maheshwari, Aishwarya~N. Reganti, Samiksha Gupta, Anupam Jamatia,
  Upendra Kumar, Björn Gambäck, and Amitava Das. 2017.
\newblock \href {https://aclanthology.org/E17-1069} {A {Societal} {Sentiment}
  {Analysis}: {Predicting} the {Values} and {Ethics} of {Individuals} by
  {Analysing} {Social} {Media} {Content}}.
\newblock In \emph{Proceedings of the 15th {Conference} of the {European}
  {Chapter} of the {Association} for {Computational} {Linguistics}: {Volume} 1,
  {Long} {Papers}}, pages 731--741, Valencia, Spain. Association for
  Computational Linguistics.

\bibitem[{Majumder et~al.(2019)Majumder, Li, Ni, and
  McAuley}]{majumderGenerating2019}
Bodhisattwa~Prasad Majumder, Shuyang Li, Jianmo Ni, and Julian McAuley. 2019.
\newblock \href {https://doi.org/10.18653/v1/D19-1613} {Generating personalized
  recipes from historical user preferences}.
\newblock In \emph{Proceedings of the 2019 conference on empirical methods in
  natural language processing and the 9th international joint conference on
  natural language processing ({EMNLP}-{IJCNLP})}, pages 5976--5982, Hong Kong,
  China. Association for Computational Linguistics.

\bibitem[{Martin~Jr. et~al.(2020)Martin~Jr., Prabhakaran, Kuhlberg, Smart, and
  Isaac}]{martinjrParticipatoryProblemFormulation2020}
Donald Martin~Jr., Vinodkumar Prabhakaran, Jill Kuhlberg, Andrew Smart, and
  William~S. Isaac. 2020.
\newblock \href {https://doi.org/10.48550/arXiv.2005.07572} {Participatory
  {Problem} {Formulation} for {Fairer} {Machine} {Learning} {Through}
  {Community} {Based} {System} {Dynamics}}.
\newblock {arXiv}:2005.07572 [cs, stat].

\bibitem[{Menick et~al.(2022)Menick, Trebacz, Mikulik, Aslanides, Song,
  Chadwick, Glaese, Young, Campbell-Gillingham, Irving, and
  McAleese}]{menickTeaching2022}
Jacob Menick, Maja Trebacz, Vladimir Mikulik, John Aslanides, Francis Song,
  Martin Chadwick, Mia Glaese, Susannah Young, Lucy Campbell-Gillingham,
  Geoffrey Irving, and Nat McAleese. 2022.
\newblock \href {https://doi.org/10.48550/arXiv.2203.11147} {Teaching language
  models to support answers with verified quotes}.
\newblock {{arXiv}}:2203.11147 [cs].

\bibitem[{Mirkin and Meunier(2015)}]{mirkinPersonalized2015}
Shachar Mirkin and Jean-Luc Meunier. 2015.
\newblock \href {https://doi.org/10.18653/v1/D15-1238} {Personalized machine
  translation: {Predicting} translational preferences}.
\newblock In \emph{Proceedings of the 2015 conference on empirical methods in
  natural language processing}, pages 2019--2025, Lisbon, Portugal. Association
  for Computational Linguistics.

\bibitem[{Mirkin et~al.(2015)Mirkin, Nowson, Brun, and
  Perez}]{mirkinMotivating2015}
Shachar Mirkin, Scott Nowson, Caroline Brun, and Julien Perez. 2015.
\newblock \href {https://doi.org/10.18653/v1/D15-1130} {Motivating
  {Personality}-aware {Machine} {Translation}}.
\newblock In \emph{Proceedings of the 2015 {Conference} on {Empirical}
  {Methods} in {Natural} {Language} {Processing}}, pages 1102--1108, Lisbon,
  Portugal. Association for Computational Linguistics.

\bibitem[{Mirzakhmedova et~al.(2023)Mirzakhmedova, Kiesel, Alshomary, Heinrich,
  Handke, Cai, Valentin, Dastgheib, Ghahroodi, Sadraei, Asgari, Kawaletz,
  Wachsmuth, and Stein}]{mirzakhmedovaTouche23ValueEval2023}
Nailia Mirzakhmedova, Johannes Kiesel, Milad Alshomary, Maximilian Heinrich,
  Nicolas Handke, Xiaoni Cai, Barriere Valentin, Doratossadat Dastgheib, Omid
  Ghahroodi, Mohammad~Ali Sadraei, Ehsaneddin Asgari, Lea Kawaletz, Henning
  Wachsmuth, and Benno Stein. 2023.
\newblock \href {http://arxiv.org/abs/2301.13771v1} {The touché23-{ValueEval}
  dataset for identifying human values behind arguments}.
\newblock {{arXiv}}: 2301.13771v1 [cs.CL].

\bibitem[{Mo et~al.(2016)Mo, Li, Zhang, Li, and Yang}]{moPersonalizing2016}
Kaixiang Mo, Shuangyin Li, Yu~Zhang, Jiajun Li, and Qiang Yang. 2016.
\newblock \href {http://arxiv.org/abs/1610.02891v3} {Personalizing a dialogue
  system with transfer reinforcement learning}.
\newblock {{arXiv}}: 1610.02891v3 [cs.AI].

\bibitem[{Nadeem et~al.(2021)Nadeem, Bethke, and
  Reddy}]{nadeemStereoSetMeasuringStereotypical2021}
Moin Nadeem, Anna Bethke, and Siva Reddy. 2021.
\newblock \href {https://doi.org/10.18653/v1/2021.acl-long.416} {{{StereoSet}}:
  {{Measuring}} stereotypical bias in pretrained language models}.
\newblock In \emph{Proceedings of the 59th {{Annual Meeting}} of the
  {{Association}} for {{Computational Linguistics}} and the 11th
  {{International Joint Conference}} on {{Natural Language Processing}}
  ({{Volume}} 1: {{Long Papers}})}, pages 5356--5371, {Online}. {Association
  for Computational Linguistics}.

\bibitem[{Nahian et~al.(2020)Nahian, Frazier, Riedl, and
  Harrison}]{nahianLearningNormsStories2020}
Md~Sultan~Al Nahian, Spencer Frazier, Mark Riedl, and Brent Harrison. 2020.
\newblock Learning norms from stories: A prior for value aligned agents.
\newblock In \emph{Proceedings of the AAAI/ACM Conference on AI, Ethics, and
  Society}, pages 124--130.

\bibitem[{Nakano et~al.(2021)Nakano, Hilton, Balaji, Wu, Ouyang, Kim, Hesse,
  Jain, Kosaraju, Saunders, Jiang, Cobbe, Eloundou, Krueger, Button, Knight,
  Chess, and Schulman}]{nakanoWebGPT2021}
Reiichiro Nakano, Jacob Hilton, Suchir Balaji, Jeff Wu, Long Ouyang, Christina
  Kim, Christopher Hesse, Shantanu Jain, Vineet Kosaraju, William Saunders,
  Xu~Jiang, Karl Cobbe, Tyna Eloundou, Gretchen Krueger, Kevin Button, Matthew
  Knight, Benjamin Chess, and John Schulman. 2021.
\newblock \href {http://arxiv.org/abs/2112.09332v3} {{WebGPT}:
  {Browser}-assisted question-answering with human feedback}.
\newblock {{arXiv}}: 2112.09332v3.

\bibitem[{Nguyen et~al.(2022)Nguyen, Nghiem, Nguyen, Tien~Le, Sabahi, Nguyen,
  and Le}]{nguyenMake2022}
Duy-Hung Nguyen, Nguyen Viet~Dung Nghiem, Bao-Sinh Nguyen, Dung~Tien Tien~Le,
  Shahab Sabahi, Minh-Tien Nguyen, and Hung Le. 2022.
\newblock \href {https://doi.org/10.18653/v1/2022.findings-naacl.147} {Make
  {The} {Most} of {Prior} {Data}: {A} {Solution} for {Interactive} {Text}
  {Summarization} with {Preference} {Feedback}}.
\newblock In \emph{Findings of the {Association} for {Computational}
  {Linguistics}: {NAACL} 2022}, pages 1919--1930, Seattle, United States.
  Association for Computational Linguistics.

\bibitem[{Nguyen et~al.(2017)Nguyen, Daumé~III, and
  Boyd-Graber}]{nguyenReinforcement2017}
Khanh Nguyen, Hal Daumé~III, and Jordan Boyd-Graber. 2017.
\newblock \href {https://doi.org/10.18653/v1/D17-1153} {Reinforcement
  {Learning} for {Bandit} {Neural} {Machine} {Translation} with {Simulated}
  {Human} {Feedback}}.
\newblock In \emph{Proceedings of the 2017 {Conference} on {Empirical}
  {Methods} in {Natural} {Language} {Processing}}, pages 1464--1474,
  Copenhagen, Denmark. Association for Computational Linguistics.

\bibitem[{Nie et~al.(2020)Nie, Zhou, and Bansal}]{nie2020can}
Yixin Nie, Xiang Zhou, and Mohit Bansal. 2020.
\newblock What can we learn from collective human opinions on natural language
  inference data?
\newblock In \emph{Proceedings of the 2020 Conference on Empirical Methods in
  Natural Language Processing (EMNLP)}, pages 9131--9143.

\bibitem[{Ouyang et~al.(2022)Ouyang, Wu, Jiang, Almeida, Wainwright, Mishkin,
  Zhang, Agarwal, Slama, Ray, Schulman, Hilton, Kelton, Miller, Simens, Askell,
  Welinder, Christiano, Leike, and Lowe}]{ouyangTraining2022}
Long Ouyang, Jeff Wu, Xu~Jiang, Diogo Almeida, Carroll~L. Wainwright, Pamela
  Mishkin, Chong Zhang, Sandhini Agarwal, Katarina Slama, Alex Ray, John
  Schulman, Jacob Hilton, Fraser Kelton, Luke Miller, Maddie Simens, Amanda
  Askell, Peter Welinder, Paul Christiano, Jan Leike, and Ryan Lowe. 2022.
\newblock \href {http://arxiv.org/abs/2203.02155v1} {Training language models
  to follow instructions with human feedback}.
\newblock {{arXiv}}: 2203.02155v1.

\bibitem[{Peng et~al.(2020)Peng, Li, Frazier, and Riedl}]{pengReducing2020}
Xiangyu Peng, Siyan Li, Spencer Frazier, and Mark Riedl. 2020.
\newblock \href {https://aclanthology.org/2020.inlg-1.43} {Reducing
  {Non}-{Normative} {Text} {Generation} from {Language} {Models}}.
\newblock In \emph{Proceedings of the 13th {International} {Conference} on
  {Natural} {Language} {Generation}}, pages 374--383, Dublin, Ireland.
  Association for Computational Linguistics.

\bibitem[{Perez et~al.(2022)Perez, Ringer, Lukošiūtė, Nguyen, Chen, Heiner,
  Pettit, Olsson, Kundu, Kadavath, Jones, Chen, Mann, Israel, Seethor,
  McKinnon, Olah, Yan, Amodei, Amodei, Drain, Li, Tran-Johnson, Khundadze,
  Kernion, Landis, Kerr, Mueller, Hyun, Landau, Ndousse, Goldberg, Lovitt,
  Lucas, Sellitto, Zhang, Kingsland, Elhage, Joseph, Mercado, DasSarma, Rausch,
  Larson, McCandlish, Johnston, Kravec, Showk, Lanham, Telleen-Lawton, Brown,
  Henighan, Hume, Bai, Hatfield-Dodds, Clark, Bowman, Askell, Grosse,
  Hernandez, Ganguli, Hubinger, Schiefer, and Kaplan}]{perezDiscovering2022}
Ethan Perez, Sam Ringer, Kamilė Lukošiūtė, Karina Nguyen, Edwin Chen, Scott
  Heiner, Craig Pettit, Catherine Olsson, Sandipan Kundu, Saurav Kadavath, Andy
  Jones, Anna Chen, Ben Mann, Brian Israel, Bryan Seethor, Cameron McKinnon,
  Christopher Olah, Da~Yan, Daniela Amodei, Dario Amodei, Dawn Drain, Dustin
  Li, Eli Tran-Johnson, Guro Khundadze, Jackson Kernion, James Landis, Jamie
  Kerr, Jared Mueller, Jeeyoon Hyun, Joshua Landau, Kamal Ndousse, Landon
  Goldberg, Liane Lovitt, Martin Lucas, Michael Sellitto, Miranda Zhang, Neerav
  Kingsland, Nelson Elhage, Nicholas Joseph, Noemí Mercado, Nova DasSarma,
  Oliver Rausch, Robin Larson, Sam McCandlish, Scott Johnston, Shauna Kravec,
  Sheer~El Showk, Tamera Lanham, Timothy Telleen-Lawton, Tom Brown, Tom
  Henighan, Tristan Hume, Yuntao Bai, Zac Hatfield-Dodds, Jack Clark, Samuel~R.
  Bowman, Amanda Askell, Roger Grosse, Danny Hernandez, Deep Ganguli, Evan
  Hubinger, Nicholas Schiefer, and Jared Kaplan. 2022.
\newblock \href {https://doi.org/10.48550/arXiv.2212.09251} {Discovering
  {Language} {Model} {Behaviors} with {Model}-{Written} {Evaluations}}.
\newblock {{arXiv}}:2212.09251 [cs].

\bibitem[{Peters et~al.(2018)Peters, Neumann, Iyyer, Gardner, Clark, Lee, and
  Zettlemoyer}]{peters2018deep}
ME~Peters, M~Neumann, M~Iyyer, M~Gardner, C~Clark, K~Lee, and L~Zettlemoyer.
  2018.
\newblock Deep contextualized word representations. arxiv 2018.
\newblock \emph{arXiv preprint {arXiv}:1802.05365}, 12.

\bibitem[{Prabhakaran et~al.(2021)Prabhakaran, Davani, and
  Diaz}]{prabhakaran2021releasing}
Vinodkumar Prabhakaran, Aida~Mostafazadeh Davani, and Mark Diaz. 2021.
\newblock On releasing annotator-level labels and information in datasets.
\newblock In \emph{Proceedings of The Joint 15th Linguistic Annotation Workshop
  (LAW) and 3rd Designing Meaning Representations (DMR) Workshop}, pages
  133--138.

\bibitem[{Preniqi et~al.(2022)Preniqi, Kalimeri, and Saitis}]{preniqiMore2022}
Vjosa Preniqi, Kyriaki Kalimeri, and Charalampos Saitis. 2022.
\newblock \href {http://arxiv.org/abs/2209.01169v1} {"{More} {Than} {Words}":
  {Linking} {Music} {Preferences} and {Moral} {Values} {Through} {Lyrics}}.
\newblock {{arXiv}}: 2209.01169v1.

\bibitem[{Pyatkin et~al.(2022)Pyatkin, Hwang, Srikumar, Lu, Jiang, Choi, and
  Bhagavatula}]{pyatkinReinforced2022}
Valentina Pyatkin, Jena~D. Hwang, Vivek Srikumar, Ximing Lu, Liwei Jiang, Yejin
  Choi, and Chandra Bhagavatula. 2022.
\newblock \href {http://arxiv.org/abs/2212.10409v1} {Reinforced clarification
  question generation with defeasibility rewards for disambiguating social and
  moral situations}.
\newblock {{arXiv}}: 2212.10409v1 [cs.CL].

\bibitem[{Qiu et~al.(2021)Qiu, Zhao, Li, Lu, Peng, Gao, and
  Zhu}]{qiuValueNet2021}
Liang Qiu, Yizhou Zhao, Jinchao Li, Pan Lu, Baolin Peng, Jianfeng Gao, and
  Song-Chun Zhu. 2021.
\newblock \href {https://doi.org/10.1609/aaai.v36i10.21368} {{ValueNet}: {A}
  {New} {Dataset} for {Human} {Value} {Driven} {Dialogue} {System}}.
\newblock {{arXiv}}: 2112.06346v1.

\bibitem[{Rabinovich et~al.(2017)Rabinovich, Patel, Mirkin, Specia, and
  Wintner}]{rabinovichPersonalized2017a}
Ella Rabinovich, Raj~Nath Patel, Shachar Mirkin, Lucia Specia, and Shuly
  Wintner. 2017.
\newblock \href {https://aclanthology.org/E17-1101} {Personalized {Machine}
  {Translation}: {Preserving} {Original} {Author} {Traits}}.
\newblock In \emph{Proceedings of the 15th {Conference} of the {European}
  {Chapter} of the {Association} for {Computational} {Linguistics}: {Volume} 1,
  {Long} {Papers}}, pages 1074--1084, Valencia, Spain. Association for
  Computational Linguistics.

\bibitem[{Rafailov et~al.(2023)Rafailov, Sharma, Mitchell, Ermon, Manning, and
  Finn}]{rafailovDirect2023}
Rafael Rafailov, Archit Sharma, Eric Mitchell, Stefano Ermon, Christopher~D.
  Manning, and Chelsea Finn. 2023.
\newblock \href {https://doi.org/10.48550/arXiv.2305.18290} {Direct
  {{Preference Optimization}}: {{Your Language Model}} is {{Secretly}} a
  {{Reward Model}}}.
\newblock {arXiv}: 2305.18290.

\bibitem[{Rashkin et~al.(2019)Rashkin, Smith, Li, and
  Boureau}]{rashkinEmpathetic2019}
Hannah Rashkin, Eric~Michael Smith, Margaret Li, and Y-Lan Boureau. 2019.
\newblock \href {https://doi.org/10.18653/v1/P19-1534} {Towards {Empathetic}
  {Open}-domain {Conversation} {Models}: {A} {New} {Benchmark} and {Dataset}}.
\newblock In \emph{Proceedings of the 57th {Annual} {Meeting} of the
  {Association} for {Computational} {Linguistics}}, pages 5370--5381, Florence,
  Italy. Association for Computational Linguistics.

\bibitem[{R{\"o}ttger et~al.(2022)R{\"o}ttger, Vidgen, Hovy, and
  Pierrehumbert}]{rottger2022two}
Paul R{\"o}ttger, Bertie Vidgen, Dirk Hovy, and Janet Pierrehumbert. 2022.
\newblock Two contrasting data annotation paradigms for subjective nlp tasks.
\newblock In \emph{Proceedings of the 2022 Conference of the North American
  Chapter of the Association for Computational Linguistics: Human Language
  Technologies}, pages 175--190.

\bibitem[{Sap et~al.(2020)Sap, Gabriel, Qin, Jurafsky, Smith, and
  Choi}]{sapSocial2020}
Maarten Sap, Saadia Gabriel, Lianhui Qin, Dan Jurafsky, Noah~A. Smith, and
  Yejin Choi. 2020.
\newblock \href {https://doi.org/10.18653/v1/2020.acl-main.486} {Social {Bias}
  {Frames}: {Reasoning} about {Social} and {Power} {Implications} of
  {Language}}.
\newblock In \emph{Proceedings of the 58th {Annual} {Meeting} of the
  {Association} for {Computational} {Linguistics}}, pages 5477--5490, Online.
  Association for Computational Linguistics.

\bibitem[{Sap et~al.(2019)Sap, Rashkin, Chen, Le~Bras, and
  Choi}]{sapSocial2019}
Maarten Sap, Hannah Rashkin, Derek Chen, Ronan Le~Bras, and Yejin Choi. 2019.
\newblock \href {https://doi.org/10.18653/v1/D19-1454} {Social {IQa}:
  {Commonsense} {Reasoning} about {Social} {Interactions}}.
\newblock In \emph{Proceedings of the 2019 {Conference} on {Empirical}
  {Methods} in {Natural} {Language} {Processing} and the 9th {International}
  {Joint} {Conference} on {Natural} {Language} {Processing}
  ({EMNLP}-{IJCNLP})}, pages 4463--4473, Hong Kong, China. Association for
  Computational Linguistics.

\bibitem[{Scheurer et~al.(2022)Scheurer, Campos, Chan, Chen, Cho, and
  Perez}]{scheurerTraining2022}
Jérémy Scheurer, Jon~Ander Campos, Jun~Shern Chan, Angelica Chen, Kyunghyun
  Cho, and Ethan Perez. 2022.
\newblock \href {https://doi.org/10.48550/arXiv.2204.14146} {Training
  {Language} {Models} with {Language} {Feedback}}.
\newblock {{arXiv}}:2204.14146 [cs].

\bibitem[{Schramowski et~al.(2019)Schramowski, Turan, Jentzsch, Rothkopf, and
  Kersting}]{schramowskiBERT2019}
Patrick Schramowski, Cigdem Turan, Sophie Jentzsch, Constantin Rothkopf, and
  Kristian Kersting. 2019.
\newblock \href {http://arxiv.org/abs/1912.05238v1} {{BERT} has a {Moral}
  {Compass}: {Improvements} of ethical and moral values of machines}.
\newblock {{arXiv}}: 1912.05238v1.

\bibitem[{Seminck and Amsili(2017)}]{seminckComputational2017}
Olga Seminck and Pascal Amsili. 2017.
\newblock \href {https://aclanthology.org/E17-4006} {A {Computational} {Model}
  of {Human} {Preferences} for {Pronoun} {Resolution}}.
\newblock In \emph{Proceedings of the {Student} {Research} {Workshop} at the
  15th {Conference} of the {European} {Chapter} of the {Association} for
  {Computational} {Linguistics}}, pages 53--63, Valencia, Spain. Association
  for Computational Linguistics.

\bibitem[{Smith et~al.(2022)Smith, Kambadur, Presani, and
  Williams}]{smithSorryHearThat2022}
Eric~Michael Smith, Melissa Hall~Melanie Kambadur, Eleonora Presani, and Adina
  Williams. 2022.
\newblock \href {https://doi.org/10.48550/arXiv.2205.09209} {"{{I}}'m sorry to
  hear that": Finding bias in language models with a holistic descriptor
  dataset}.

\bibitem[{Solaiman and Dennison(2021)}]{solaimanProcess2021}
Irene Solaiman and Christy Dennison. 2021.
\newblock \href
  {https://proceedings.neurips.cc/paper/2021/hash/2e855f9489df0712b4bd8ea9e2848c5a-Abstract.html}
  {Process for {Adapting} {Language} {Models} to {Society} ({PALMS}) with
  {Values}-{Targeted} {Datasets}}.
\newblock In \emph{Advances in {Neural} {Information} {Processing} {Systems}},
  volume~34, pages 5861--5873. Curran Associates, Inc.

\bibitem[{Song et~al.(2023)Song, Yu, Li, Yu, Huang, Li, and
  Wang}]{songPreference2023}
Feifan Song, Bowen Yu, Minghao Li, Haiyang Yu, Fei Huang, Yongbin Li, and
  Houfeng Wang. 2023.
\newblock \href {https://doi.org/10.48550/arXiv.2306.17492} {Preference
  {{Ranking Optimization}} for {{Human Alignment}}}.
\newblock {arXiv}: 2306.17492.

\bibitem[{Stiennon et~al.(2020)Stiennon, Ouyang, Wu, Ziegler, Lowe, Voss,
  Radford, Amodei, and Christiano}]{stiennonLearning2020}
Nisan Stiennon, Long Ouyang, Jeff Wu, Daniel~M. Ziegler, Ryan Lowe, Chelsea
  Voss, Alec Radford, Dario Amodei, and Paul Christiano. 2020.
\newblock \href {http://arxiv.org/abs/2009.01325v3} {Learning to summarize from
  human feedback}.
\newblock {{arXiv}}: 2009.01325v3.

\bibitem[{Tay et~al.(2020)Tay, Ong, Fu, Chan, Chen, Luu, and
  Pal}]{tayWould2020}
Yi~Tay, Donovan Ong, Jie Fu, Alvin Chan, Nancy Chen, Anh~Tuan Luu, and Chris
  Pal. 2020.
\newblock \href {https://doi.org/10.18653/v1/2020.acl-main.477} {Would you
  {Rather}? {A} {New} {Benchmark} for {Learning} {Machine} {Alignment} with
  {Cultural} {Values} and {Social} {Preferences}}.
\newblock In \emph{Proceedings of the 58th {Annual} {Meeting} of the
  {Association} for {Computational} {Linguistics}}, pages 5369--5373, Online.
  Association for Computational Linguistics.

\bibitem[{Thoppilan et~al.(2022)Thoppilan, De~Freitas, Hall, Shazeer,
  Kulshreshtha, Cheng, Jin, Bos, Baker, Du, Li, Lee, Zheng, Ghafouri, Menegali,
  Huang, Krikun, Lepikhin, Qin, Chen, Xu, Chen, Roberts, Bosma, Zhao, Zhou,
  Chang, Krivokon, Rusch, Pickett, Srinivasan, Man, Meier-Hellstern, Morris,
  Doshi, Santos, Duke, Soraker, Zevenbergen, Prabhakaran, Diaz, Hutchinson,
  Olson, Molina, Hoffman-John, Lee, Aroyo, Rajakumar, Butryna, Lamm, Kuzmina,
  Fenton, Cohen, Bernstein, Kurzweil, Aguera-Arcas, Cui, Croak, Chi, and
  Le}]{thoppilanLaMDA2022}
Romal Thoppilan, Daniel De~Freitas, Jamie Hall, Noam Shazeer, Apoorv
  Kulshreshtha, Heng-Tze Cheng, Alicia Jin, Taylor Bos, Leslie Baker, Yu~Du,
  YaGuang Li, Hongrae Lee, Huaixiu~Steven Zheng, Amin Ghafouri, Marcelo
  Menegali, Yanping Huang, Maxim Krikun, Dmitry Lepikhin, James Qin, Dehao
  Chen, Yuanzhong Xu, Zhifeng Chen, Adam Roberts, Maarten Bosma, Vincent Zhao,
  Yanqi Zhou, Chung-Ching Chang, Igor Krivokon, Will Rusch, Marc Pickett,
  Pranesh Srinivasan, Laichee Man, Kathleen Meier-Hellstern, Meredith~Ringel
  Morris, Tulsee Doshi, Renelito~Delos Santos, Toju Duke, Johnny Soraker, Ben
  Zevenbergen, Vinodkumar Prabhakaran, Mark Diaz, Ben Hutchinson, Kristen
  Olson, Alejandra Molina, Erin Hoffman-John, Josh Lee, Lora Aroyo, Ravi
  Rajakumar, Alena Butryna, Matthew Lamm, Viktoriya Kuzmina, Joe Fenton, Aaron
  Cohen, Rachel Bernstein, Ray Kurzweil, Blaise Aguera-Arcas, Claire Cui,
  Marian Croak, Ed~Chi, and Quoc Le. 2022.
\newblock \href {https://doi.org/10.48550/arXiv.2201.08239} {{LaMDA}:
  {Language} {Models} for {Dialog} {Applications}}.
\newblock {{arXiv}}:2201.08239 [cs].

\bibitem[{Tversky(1969)}]{tversky1969intransitivity}
Amos Tversky. 1969.
\newblock Intransitivity of preferences.
\newblock \emph{Psychological review}, 76(1):31.

\bibitem[{Vaswani et~al.(2017)Vaswani, Shazeer, Parmar, Uszkoreit, Jones,
  Gomez, Kaiser, and Polosukhin}]{vaswaniAttentionAllYou2017}
Ashish Vaswani, Noam Shazeer, Niki Parmar, Jakob Uszkoreit, Llion Jones,
  Aidan~N. Gomez, Lukasz Kaiser, and Illia Polosukhin. 2017.
\newblock \href {https://doi.org/10.48550/arXiv.1706.03762} {Attention {{Is All
  You Need}}}.

\bibitem[{Wang et~al.(2021)Wang, Wei, Zhang, Huang, Xie, and
  Chen}]{wangNonParametric2021}
Dongqi Wang, Haoran Wei, Zhirui Zhang, Shujian Huang, Jun Xie, and Jiajun Chen.
  2021.
\newblock \href {http://arxiv.org/abs/2109.11136v3} {Non-{Parametric} {Online}
  {Learning} from {Human} {Feedback} for {Neural} {Machine} {Translation}}.
\newblock {{arXiv}}: 2109.11136v3.

\bibitem[{Wang et~al.(2017)Wang, Wang, Liu, Wang, Wang, and
  Wang}]{wangPredicting2017}
Xin Wang, Jianan Wang, Yuanchao Liu, Xiaolong Wang, Zhuoran Wang, and Baoxun
  Wang. 2017.
\newblock \href {https://aclanthology.org/I17-1072} {Predicting {Users}'
  {Negative} {Feedbacks} in {Multi}-{Turn} {Human}-{Computer} {Dialogues}}.
\newblock In \emph{Proceedings of the {Eighth} {International} {Joint}
  {Conference} on {Natural} {Language} {Processing} ({Volume} 1: {Long}
  {Papers})}, pages 713--722, Taipei, Taiwan. Asian Federation of Natural
  Language Processing.

\bibitem[{Wang et~al.(2022)Wang, Kordi, Mishra, Liu, Smith, Khashabi, and
  Hajishirzi}]{wangSelfInstruct2022}
Yizhong Wang, Yeganeh Kordi, Swaroop Mishra, Alisa Liu, Noah~A. Smith, Daniel
  Khashabi, and Hannaneh Hajishirzi. 2022.
\newblock \href {http://arxiv.org/abs/2212.10560v1} {Self-{Instruct}:
  {Aligning} {Language} {Model} with {Self} {Generated} {Instructions}}.
\newblock {{arXiv}}: 2212.10560v1.

\bibitem[{Welbl et~al.(2021)Welbl, Glaese, Uesato, Dathathri, Mellor,
  Hendricks, Anderson, Kohli, Coppin, and
  Huang}]{welblChallengesDetoxifyingLanguage2021}
Johannes Welbl, Amelia Glaese, Jonathan Uesato, Sumanth Dathathri, John Mellor,
  Lisa~Anne Hendricks, Kirsty Anderson, Pushmeet Kohli, Ben Coppin, and Po-Sen
  Huang. 2021.
\newblock \href {https://doi.org/10.48550/arXiv.2109.07445} {Challenges in
  {{Detoxifying Language Models}}}.

\bibitem[{Westland(2022)}]{westland2022information}
J~Christopher Westland. 2022.
\newblock Information loss and bias in likert survey responses.
\newblock \emph{Plos one}, 17(7):e0271949.

\bibitem[{Wu et~al.(2021)Wu, Ouyang, Ziegler, Stiennon, Lowe, Leike, and
  Christiano}]{wuRecursively2021}
Jeff Wu, Long Ouyang, Daniel~M. Ziegler, Nisan Stiennon, Ryan Lowe, Jan Leike,
  and Paul Christiano. 2021.
\newblock \href {http://arxiv.org/abs/2109.10862v2} {Recursively {Summarizing}
  {Books} with {Human} {Feedback}}.
\newblock {{arXiv}}: 2109.10862v2.

\bibitem[{Wu et~al.(2023)Wu, Hu, Shi, Dziri, Suhr, Ammanabrolu, Smith,
  Ostendorf, and Hajishirzi}]{wuFineGrained2023}
Zeqiu Wu, Yushi Hu, Weijia Shi, Nouha Dziri, Alane Suhr, Prithviraj
  Ammanabrolu, Noah~A. Smith, Mari Ostendorf, and Hannaneh Hajishirzi. 2023.
\newblock \href {https://doi.org/10.48550/arXiv.2306.01693} {Fine-{{Grained
  Human Feedback Gives Better Rewards}} for {{Language Model Training}}}.
\newblock {arXiv}: 2306.01693.

\bibitem[{Xu et~al.(2021{\natexlab{a}})Xu, Ju, Li, Boureau, Weston, and
  Dinan}]{xuBotAdversarial2021}
Jing Xu, Da~Ju, Margaret Li, Y-Lan Boureau, Jason Weston, and Emily Dinan.
  2021{\natexlab{a}}.
\newblock \href {https://doi.org/10.18653/v1/2021.naacl-main.235}
  {Bot-{Adversarial} {Dialogue} for {Safe} {Conversational} {Agents}}.
\newblock In \emph{Proceedings of the 2021 {Conference} of the {North}
  {American} {Chapter} of the {Association} for {Computational} {Linguistics}:
  {Human} {Language} {Technologies}}, pages 2950--2968, Online. Association for
  Computational Linguistics.

\bibitem[{Xu et~al.(2021{\natexlab{b}})Xu, Ju, Li, Boureau, Weston, and
  Dinan}]{xuRecipes2021}
Jing Xu, Da~Ju, Margaret Li, Y.-Lan Boureau, Jason Weston, and Emily Dinan.
  2021{\natexlab{b}}.
\newblock \href {https://doi.org/10.48550/arXiv.2010.07079} {Recipes for
  {Safety} in {Open}-domain {Chatbots}}.
\newblock {{arXiv}}:2010.07079 [cs].

\bibitem[{Xu et~al.(2022)Xu, Ung, Komeili, Arora, Boureau, and
  Weston}]{xuLearning2022}
Jing Xu, Megan Ung, Mojtaba Komeili, Kushal Arora, Y.-Lan Boureau, and Jason
  Weston. 2022.
\newblock \href {http://arxiv.org/abs/2208.03270v2} {Learning {New} {Skills}
  after {Deployment}: {Improving} open-domain internet-driven dialogue with
  human feedback}.
\newblock {{arXiv}}: 2208.03270v2.

\bibitem[{Yuan et~al.(2023)Yuan, Yuan, Tan, Wang, Huang, and
  Huang}]{yuanRRHF2023}
Zheng Yuan, Hongyi Yuan, Chuanqi Tan, Wei Wang, Songfang Huang, and Fei Huang.
  2023.
\newblock \href {https://doi.org/10.48550/arXiv.2304.05302} {{{RRHF}}: {{Rank
  Responses}} to {{Align Language Models}} with {{Human Feedback}} without
  tears}.
\newblock {arXiv}: 2304.05302.

\bibitem[{Zhang et~al.(2018)Zhang, Dinan, Urbanek, Szlam, Kiela, and
  Weston}]{zhangPersonalizing2018}
Saizheng Zhang, Emily Dinan, Jack Urbanek, Arthur Szlam, Douwe Kiela, and Jason
  Weston. 2018.
\newblock \href {https://doi.org/10.18653/v1/P18-1205} {Personalizing
  {Dialogue} {Agents}: {I} have a dog, do you have pets too?}
\newblock In \emph{Proceedings of the 56th {Annual} {Meeting} of the
  {Association} for {Computational} {Linguistics} ({Volume} 1: {Long}
  {Papers})}, pages 2204--2213, Melbourne, Australia. Association for
  Computational Linguistics.

\bibitem[{Zhao et~al.(2021)Zhao, Khashabi, Khot, Sabharwal, and
  Chang}]{zhaoEthicalAdvice2021}
Jieyu Zhao, Daniel Khashabi, Tushar Khot, Ashish Sabharwal, and Kai-Wei Chang.
  2021.
\newblock \href {https://doi.org/10.18653/v1/2021.findings-acl.364}
  {Ethical-{Advice} {Taker}: {Do} {Language} {Models} {Understand} {Natural}
  {Language} {Interventions}?}
\newblock In \emph{Findings of the {Association} for {Computational}
  {Linguistics}: {ACL}-{IJCNLP} 2021}, pages 4158--4164, Online. Association
  for Computational Linguistics.

\bibitem[{Zhou et~al.(2023)Zhou, Liu, Xu, Iyer, Sun, Mao, Ma, Efrat, Yu, Yu,
  Zhang, Ghosh, Lewis, Zettlemoyer, and Levy}]{zhouLIMA2023}
Chunting Zhou, Pengfei Liu, Puxin Xu, Srini Iyer, Jiao Sun, Yuning Mao, Xuezhe
  Ma, Avia Efrat, Ping Yu, Lili Yu, Susan Zhang, Gargi Ghosh, Mike Lewis, Luke
  Zettlemoyer, and Omer Levy. 2023.
\newblock \href {http://arxiv.org/abs/2305.11206} {{{LIMA}}: {{Less Is More}}
  for {{Alignment}}}.
\newblock {arXiv}: 2305.11206.

\bibitem[{Zhou et~al.(2021)Zhou, Deshmukh, Greer, and Lee}]{zhouNaRLE2021}
Ruijie Zhou, Soham Deshmukh, Jeremiah Greer, and Charles Lee. 2021.
\newblock \href {http://arxiv.org/abs/2110.02148v1} {{NaRLE}: {Natural}
  language models using reinforcement learning with emotion feedback}.
\newblock {{arXiv}}: 2110.02148v1 [cs.CL].

\bibitem[{Ziegler et~al.(2019)Ziegler, Stiennon, Wu, Brown, Radford, Amodei,
  Christiano, and Irving}]{zieglerFineTuning2019}
Daniel~M. Ziegler, Nisan Stiennon, Jeffrey Wu, Tom~B. Brown, Alec Radford,
  Dario Amodei, Paul Christiano, and Geoffrey Irving. 2019.
\newblock \href {http://arxiv.org/abs/1909.08593v2} {Fine-{Tuning} {Language}
  {Models} from {Human} {Preferences}}.
\newblock {{arXiv}}: 1909.08593v2.

\bibitem[{Ziems et~al.(2022)Ziems, Yu, Wang, Halevy, and Yang}]{ziemsMoral2022}
Caleb Ziems, Jane Yu, Yi-Chia Wang, Alon Halevy, and Diyi Yang. 2022.
\newblock \href {https://doi.org/10.18653/v1/2022.acl-long.261} {The {Moral}
  {Integrity} {Corpus}: {A} {Benchmark} for {Ethical} {Dialogue} {Systems}}.
\newblock In \emph{Proceedings of the 60th {Annual} {Meeting} of the
  {Association} for {Computational} {Linguistics} ({Volume} 1: {Long}
  {Papers})}, pages 3755--3773, Dublin, Ireland. Association for Computational
  Linguistics.

\bibitem[{Zytko et~al.(2022)Zytko, J.~Wisniewski, Guha, P.~S.~Baumer, and
  Lee}]{zytkoParticipatory2022}
Douglas Zytko, Pamela J.~Wisniewski, Shion Guha, Eric P.~S.~Baumer, and
  Min~Kyung Lee. 2022.
\newblock \href {https://doi.org/10.1145/3491101.3516506} {Participatory
  {Design} of {AI} {Systems}: {Opportunities} and {Challenges} {Across}
  {Diverse} {Users}, {Relationships}, and {Application} {Domains}}.
\newblock In \emph{Extended {Abstracts} of the 2022 {CHI} {Conference} on
  {Human} {Factors} in {Computing} {Systems}}, {CHI} {EA} '22, pages 1--4, New
  York, NY, USA. Association for Computing Machinery.

\end{thebibliography}
\bibliographystyle{acl_natbib}

\appendix
\section{Flowchart of Articles for Scoping the Review}
\label{sec:flowchart}
\begin{figure*}
    \centering
    \includegraphics[width = 0.9\textwidth]{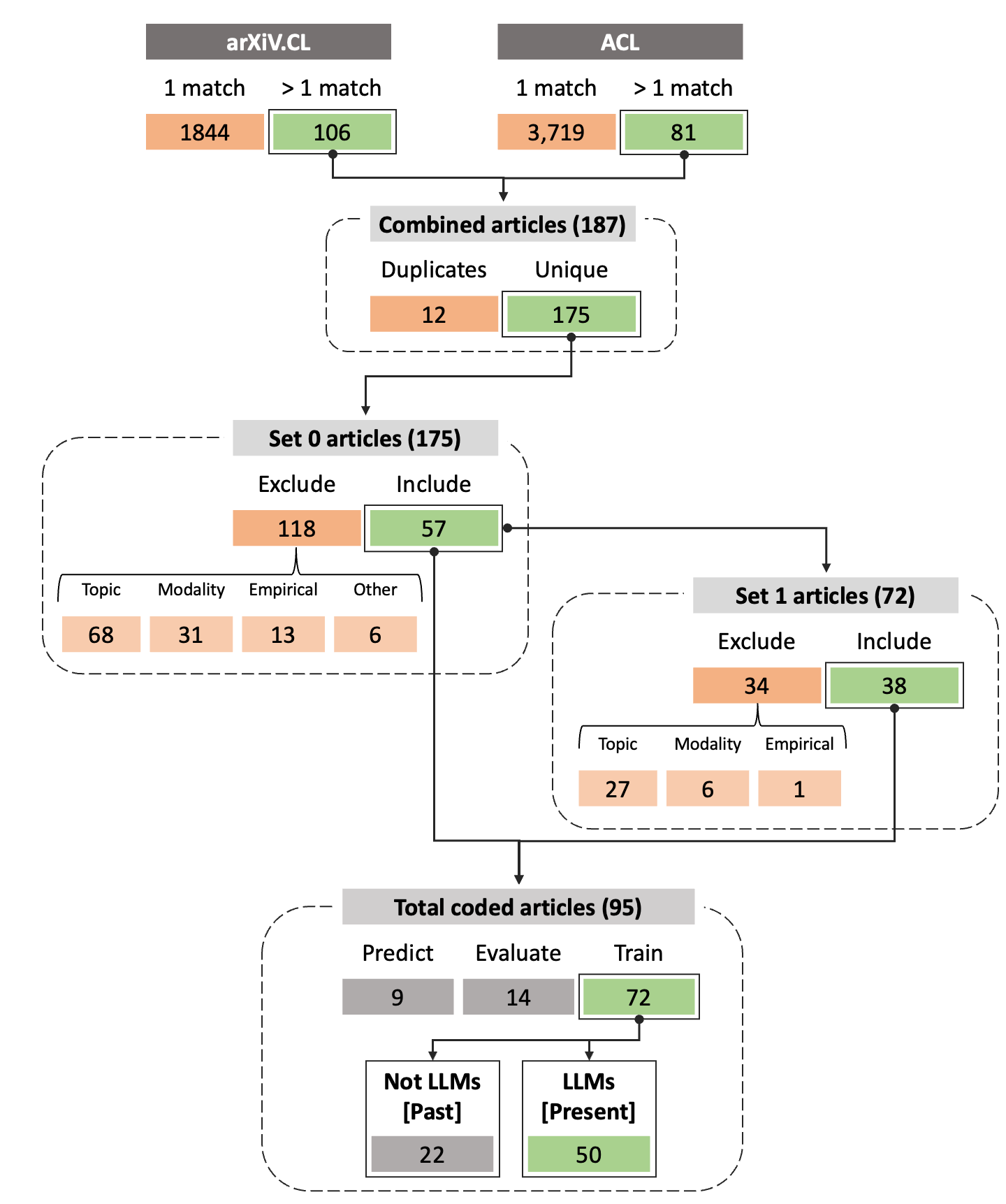}
    \caption{\textbf{Flowchart of the selection process for articles in our review.} We first match on the keywords defined in \cref{tab:keywords_tab}, keeping only articles with >1 matches in the title. We then de-duplicate articles posted on both arXiv and ACL. This initial set is called $S_0$. We apply the inclusion criteria listed in \cref{sec:selecting}, and also add any possibly relevant references to the snowballed set ($S_1$) regardless of publishing venue. We also apply the inclusion criteria to $S_1$. Finally, we make two additional categorisations --- the dominant contribution type of the article (predict, evaluate or train) and whether it uses LLMs.}
    \label{fig:flowchart}
\end{figure*}

In \cref{fig:flowchart}, we schematically summarise the process of selecting articles for our review.

\begin{table*}[h!]
\centering
\small
\renewcommand{\arraystretch}{1.75}
\setlength\tabcolsep{1.5pt}
\resizebox{\textwidth}{!}{  
\begin{tabular}{m{0.25\textwidth}m{0.7\textwidth}}
\hline
\rowcolor[gray]{0.9}
\textbf{Task} & \textbf{References} \\
\hline
\rowcolor[gray]{0.99}
Text generation & \cite{pengReducing2020,  liuMitigating2021, solaimanProcess2021,  aroraDirector2022,liuAligning2022, korbakPretraining2023}, \textit{including story generation} \cite{castricatoRobust2022} and \textit{code generation} \cite{korbakPretraining2023} \\
\rowcolor[gray]{0.8}
Instruction following & \cite{honovichUnnatural2022, ouyangTraining2022, wangSelfInstruct2022} \\
\rowcolor[gray]{0.99}
Open-ended dialogue & \cite{hancockLearning2019, gaoDialogue2020, jaquesHumancentric2020, askellGeneral2021, qiuValueNet2021, thoppilanLaMDA2022, baiConstitutional2022, baiTraining2022, ganguliRed2022, luBoosting2022, xuLearning2022, liuLanguages2023}, \textit{including information-seeking dialogue} \cite{glaeseImproving2022}\\
\rowcolor[gray]{0.8}
Open-book generative QA & \cite{zhaoEthicalAdvice2021, dengPersonalized2021, nakanoWebGPT2021, menickTeaching2022} \\
\rowcolor[gray]{0.99}
Summarization & \cite{gaoAPRIL2018, bohmBetter2019, zieglerFineTuning2019, stiennonLearning2020, scheurerTraining2022, nguyenMake2022,  liuLanguages2023}, \textit{including long-form book summarisation} \cite{wuRecursively2021} and \textit{opinion consensus summarisation} \cite{bakkerFinetuning2022} \\
\rowcolor[gray]{0.8}
Toxic language & \cite{dinanBuild2019, pengReducing2020, liuDExperts2021, scheurerTraining2022, juLearning2022, bangEnabling2022, liuAligning2022} \\
\rowcolor[gray]{0.99}
Moral \& normative judgements & \cite{forbesSocial2020, nahianLearningNormsStories2020, jiangCan2022,  liuAligning2022, jinWhen2022, pyatkinReinforced2022, liuSecond2023} \\
\rowcolor[gray]{0.8}
Others & \textit{Sentiment and style transfer} \cite{zieglerFineTuning2019, pengReducing2020, liuDExperts2021}; \textit{recipe generation} \cite{majumderGenerating2019}; \textit{predicting intent of emails} \cite{zhouNaRLE2021}; \textit{machine translation} \cite{wangNonParametric2021}\\
\hline
\end{tabular}
}
\caption{Articles categorised by target task.}
\label{tab:tasks}
\end{table*}

\section{Code Book}
\label{sec:code_book}
We present the full code book used for each article in \cref{tab:code_book}. These questions were inputted into an online form then coded by two authors of the paper, with frequent check-ins to ensure similarity of interpretation on how the form should be used. The first theme (conceptual) makes up our conceptual comments in the main paper, while the laboural and technical themes make up our methodological comments in the main paper. 

\section{Additional Information on Reviewed Articles}

\label{sec:additional_appendix_info}
\subsection{Target Tasks}
In \cref{tab:tasks}, we summarise the core target tasks approached by each article. Reflecting the recent movement away from specialist NLP systems towards general purpose language agents, the majority of articles work with generalised models that can handle many other NLP subtasks via instruction or dialogue. 

\subsection{Evaluating Models}
\label{sec:evaluating}
Even articles employing indirect or simulated human feedback usually conduct a human evaluation stage \cite{pengReducing2020, liuMitigating2021, liuAligning2022}. Differently-trained models are often compared via ELO scores or win rates \cite{zieglerFineTuning2019, nakanoWebGPT2021, baiConstitutional2022, baiTraining2022, scheurerTraining2022, bakkerFinetuning2022, glaeseImproving2022, ouyangTraining2022}. Most evaluations include fine-grained questions about model outputs, including quality or usefulness \cite{wuRecursively2021, nakanoWebGPT2021, liuAligning2022, bakkerFinetuning2022}; political bias \cite{liuMitigating2021}; coherence \cite{wuRecursively2021, nakanoWebGPT2021, liuAligning2022, bakkerFinetuning2022}; safety or harmlessness \cite{xuBotAdversarial2021, luBoosting2022, ganguliRed2022, thoppilanLaMDA2022}; informativeness, correctness or trustworthiness \cite{wuRecursively2021, nakanoWebGPT2021, luBoosting2022}; creativity \cite{honovichUnnatural2022}; and alignment with a human value or trait \cite{solaimanProcess2021, liuAligning2022, castricatoRobust2022, liuLanguages2023}.

Others use automated metrics to quantitatively compare models and outputs, with \citet{bohmBetter2019} and \citet{stiennonLearning2020} performing a comparison of how such automated metrics correlate with human preferences. Metrics include ROGUE \cite{bohmBetter2019,  zieglerFineTuning2019, stiennonLearning2020, liuAligning2022, nguyenMake2022, wangSelfInstruct2022, wuRecursively2021, liuSecond2023}, summary length \cite{stiennonLearning2020}, perplexity \cite{liuMitigating2021,  liuAligning2022, liuSecond2023} or SacreBLEU \cite{wangNonParametric2021}. Sometimes separate discriminative classifier are deployed to measure textual attributes \cite{thoppilanLaMDA2022}, such as toxicity measured via Perspective API scores \cite{solaimanProcess2021, aroraDirector2022}. \citet{scheurerTraining2022} score how close feedback and refinements are in the embedding space because they find written feedback often describes an ``ideal'' output. Any prediction tasks -- e.g., whether an ethical judgement is fair or unfair \cite{jiangCan2022}, a situation is normative or non-normative \cite{nahianLearningNormsStories2020, forbesSocial2020}, a norm exception is permissible or not permissible \cite{jinWhen2022} or an utterance is value aligned or misaligned \cite{qiuValueNet2021} -- use F1-score or accuracy as evaluation metrics.

Metrics or human evaluations that measure how aligned a resultant model is with human preferences or values can be contrasted with general investigations of model capabilities to estimate the so-called ``alignment tax'' \cite{liuAligning2022}. For instance, \citet{korbakPretraining2023} rely on two metrics: (i) misalignment score, calculated using the same automated reward functions as training (toxicity score, number of PII instances per character, number of PEP errors per character), and (ii) capability score, calculated as the KL divergence of output distribution from a highly capable model (GPT-3). Some articles assess the drop in other performance measures on NLP benchmark tasks measuring truthfulness, toxicity or bias \cite{baiTraining2022, ouyangTraining2022}.

\section{Articles with Other Contribution Types}
\label{sec:other_conts}
In the main paper, we discuss papers that seek to embed, train or align LLMs with human preferences and values. Here, we give a brief overview of the other categories of papers which are excluded from the main review. 

\paragraph{Predict} These articles include detecting moral content from tweets \cite{hooverMoral2020, asprinoUncovering2022} or adapting to moral shifts \cite{huangLearning2022}; predicting values and ethics from social media content \cite{maheshwariSocietal2017} or music preferences \cite{preniqiMore2022}; linking event or entity extraction with moral values in knowledge bases \cite{linAcquiring2017, liMultilingual2019}; and identifying human values in arguments \cite{kieselIdentifying2022}.

\paragraph{Evaluate} These articles include those that benchmark judgements in moral or ethical situations \cite{tayWould2020, hendrycksAligning2020, lourieSCRUPLES2021, ziemsMoral2022, mirzakhmedovaTouche23ValueEval2023}; assess social biases or social reasoning \cite{sapSocial2019, sapSocial2020}; evaluate performance on personality-aware dialogue \cite{zhangPersonalizing2018} or empathetic dialogue \cite{rashkinEmpathetic2019}; and detect non-human identity in conversations \cite{grosRUARobot2021}. Others directly evaluate the values or traits of existing models \cite{schramowskiBERT2019, jentzschSemantics2019, perezDiscovering2022}. 

\clearpage

\begin{table*}
\centering
\footnotesize
\renewcommand{\arraystretch}{1.25}
\begin{tabularx}{\textwidth}{>{\bfseries\RaggedRight}p{2.5cm}>{\RaggedRight}X>{\RaggedRight}X}
\hline
\rowcolor[rgb]{1,0.906,0.741}
\multicolumn{3}{c}{\textbf{METADATA}} \\
\hline
relevance & Whether to include or exclude article from set & single choice: [exclude, include] \\
\rowcolor[rgb]{0.898,0.898,0.898}
exclusion criteria & Reason for excluding the article & single choice: [topic, modality, empirical, other] \\
exclusion detail & Additional text summary of exclusion criteria reasoning & free-text \\
\rowcolor[rgb]{0.898,0.898,0.898}
snowball keys & The bib keys of additional references & free-text \\
contribution type & Dominant contribution type of article & single choice: [evaluate, predict, train, other] \\
\rowcolor[rgb]{0.898,0.898,0.898}
contribution detail & How are the main contributions of the article described? & free-text \\
about LLMs & Whether the article uses LLMs & single choice: [yes, no] \\
\rowcolor[rgb]{0.898,0.898,0.898}
short summary & 1-3 sentence summary of the article & free-text \\
\rowcolor[rgb]{0.741,0.867,0.741}
\hline
\multicolumn{3}{c}{\textbf{CONCEPTUAL THEME}} \\
\hline
\rowcolor[rgb]{0.898,0.898,0.898}
terminology & Is feedback discussed using the terms `preferences' or `values'? & single choice: [ preferences, values, mix, other] \\
motivation & What is the motivation for feedback learning? & free-text \\
\rowcolor[rgb]{0.898,0.898,0.898}
target concepts & Which human values or preferences are prioritized or included? & free-text \\
concept defs & How are human values or preferences defined? & free-text \\
\rowcolor[rgb]{0.898,0.898,0.898}
theories & What theories (if any) are used to define preferences/values? & free-text \\
concept scope & Are concepts defined as universal or culturally/contextually understood? & free-text \\
\rowcolor[rgb]{0.898,0.898,0.898}
interpretation freedom & What level of freedom are humans given in interpreting the in-scope target concepts, e.g., ``helpfulness''? & single choice: [prescriptive paradigm, subjective paradigm, unclear] \\
\rowcolor[rgb]{0.792,1,1}
\hline
\multicolumn{3}{c}{\textbf{LABORAL THEME}} \\
\hline
\rowcolor[rgb]{0.898,0.898,0.898}
feedback generation & How is feedback data collected? & multi choice: [human-generated explicit, human-generated implicit, model-generated, combined, other] \\
feedback types & What forms of feedback are collected? At what stage, and if for training or for evaluation? & free-text \\
\rowcolor[rgb]{0.898,0.898,0.898}
labour documentation & Is the labor force documented? & single choice: [yes, no, nan] \\
labour details & What level of documentation or what details are documented? & free-text \\
\rowcolor[rgb]{0.898,0.898,0.898}
labour force & Which human group(s) generate the feedback? & multi choice: [crowdworkers, in-house team authors, unknown] \\
labour force detail & What further detail is provided on who generates feedback? & free-text \\
\rowcolor[rgb]{0.898,0.898,0.898}
labour force size & How many humans are involved in feedback collection for training and/or evaluation? & free-text \\
\rowcolor[rgb]{0.894,0.792,0.894}
\hline
\multicolumn{3}{c}{\textbf{TECHNICAL THEME}} \\
\hline
\rowcolor[rgb]{0.898,0.898,0.898}
data size & What is the size of the feedback dataset for training and for evaluation? & free-text \\
intervention stage & When and how is feedback integrated into the model? & multi choice: [pre-training, fine-tuning, prompting, other] \\
\rowcolor[rgb]{0.898,0.898,0.898}
metrics & What metrics and which evaluation datasets are used? & free-text \\
model approach & Summarize the modeling methodology & free-text \\
\rowcolor[rgb]{1,0.741,0.741}
\hline
\multicolumn{3}{c}{\textbf{PROCEDURAL THEME}}\\
\hline
\rowcolor[rgb]{0.898,0.898,0.898}
authorship & Authorship composition of the article & single choice: [academia, industry, mixed] \\
data availability & Whether the data artifacts are publicly available & single choice: [yes, no, unclear] \\
\rowcolor[rgb]{0.898,0.898,0.898}
model availability & Whether the model artifacts are publicly available & single choice: [yes, no, unclear] \\
peer review & Whether the article is peer-reviewed & single choice: [yes, no, unclear] \\
\bottomrule
\end{tabularx}
\caption{\textbf{Code book used for each article included in the review}. We show the field name, the prompt or instruction for the coder and the type of response variable (including options if it is a single or multiple choice question). Frequent communication between two coders was established to ensure the fields were being applied consistently.}
\label{tab:code_book}
\end{table*}

\end{document}